\documentclass[11pt]{article}

\usepackage[final]{acl}

\usepackage{times}
\usepackage{latexsym}
\usepackage{bbm}

\usepackage[T1]{fontenc}

\usepackage[utf8]{inputenc}

\usepackage{microtype}
\usepackage{booktabs}
\usepackage{tcolorbox}
\usepackage{xcolor}
\usepackage{enumitem}
\usepackage{amsmath}
\usepackage{subcaption}
\usepackage{amssymb}
\usepackage{svg}
\newtcolorbox{promptbox}[1][]{
    colback=gray!8,           %
    colframe=blue!70!black,   %
    coltitle=white,
    fonttitle=\bfseries,
    fontupper=\fontsize{10pt}{12pt}\selectfont,  %
    title=#1,
    arc=4pt,
    boxrule=1pt,
    left=8pt,
    right=8pt,
    top=6pt,
    bottom=6pt,
}
\usepackage{inconsolata}

\usepackage{graphicx}

\title{Fairness or Fluency? An Investigation into Language Bias of Pairwise LLM-as-a-Judge}

\author{
 \textbf{Xiaolin Zhou\textsuperscript{1}\thanks{Equal contribution}},
 \textbf{Zheng Luo\textsuperscript{1}\footnotemark[1]},
 \textbf{Yicheng Gao\textsuperscript{1}\footnotemark[1]},
 \textbf{Qixuan Chen\textsuperscript{1}}, \\
 \textbf{Xiyang Hu\textsuperscript{2}}, 
 \textbf{Yue Zhao\textsuperscript{1}},
 \textbf{Ruishan Liu\textsuperscript{1}}
\\
 \textsuperscript{1}University of Southern California,
 \textsuperscript{2}Arizona State University
\\
\texttt{\{xzhou733, luozheng, gaoyiche, qixuanch, yue.z, ruishanl\}@usc.edu,}
\\
\texttt{xiyanghu@asu.edu}
}

\begin{document}
\maketitle

\begin{abstract}
Recent advances in Large Language Models (LLMs) have incentivized the development of LLM-as-a-judge, an application of LLMs where they are used as judges to decide the quality of a certain piece of text given a certain context. However, previous studies have demonstrated that LLM-as-a-judge can be biased towards different aspects of the judged texts, which often do not align with human preference. One of the identified biases is \textbf{language bias}, which indicates that the decision of LLM-as-a-judge can differ based on the language of the judged texts. In this paper, we study two types of language bias in pairwise LLM-as-a-judge: (1) performance disparity between languages when the judge is prompted to compare options from the same language, and (2) bias towards options written in major languages when the judge is prompted to compare options of two different languages. We find that for same-language judging, there exist significant performance disparities across language families, with European languages consistently outperforming African languages, and this bias is more pronounced in culturally-related subjects. For inter-language judging, we observe that most models favor English answers, and that this preference is influenced more by answer language than question language. Finally, we investigate whether language bias is in fact caused by low-perplexity bias, a previously identified bias of LLM-as-a-judge, and we find that while perplexity is slightly correlated with language bias, language bias cannot be fully explained by perplexity only.
\end{abstract}

\section{Introduction}

\begin{figure}[th]
    \centering
    \includegraphics[width=\linewidth]{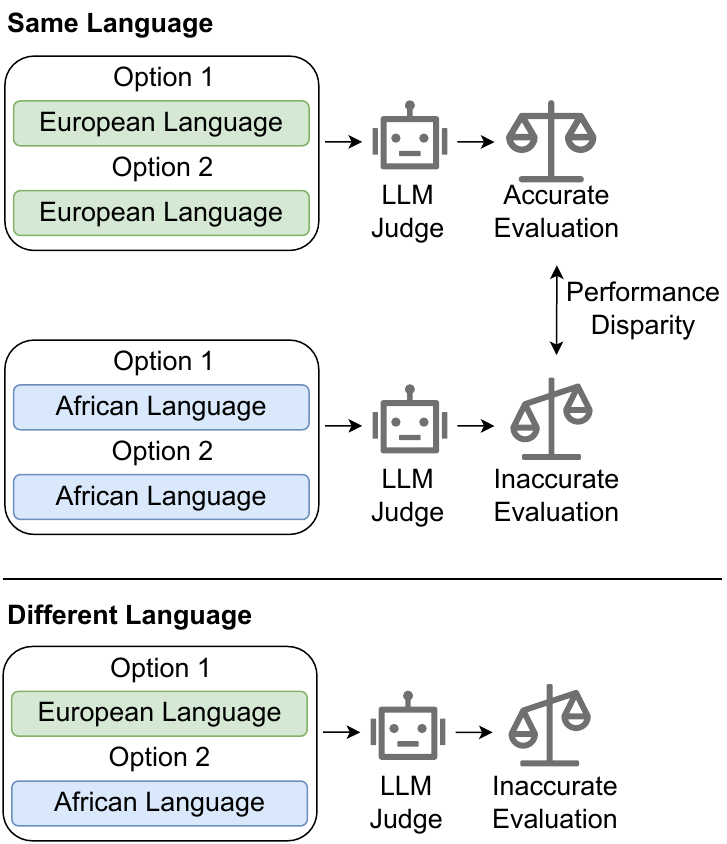}
    \caption{Illustration of language bias in the same-language judging and inter-language judging scenario.}
    \label{fig:1-pipeline}
\end{figure}

Large Language Models (LLMs) have demonstrated remarkable capabilities across a wide range of natural language processing tasks. One of the important applications of LLMs is LLM-as-a-judge. In this framework, LLMs are prompted to assess the quality of text inputs, either by assigning scores (pointwise), selecting the better option from a pair (pairwise), or ranking multiple candidates (listwise)~\citep{li2024llmsasjudgescomprehensivesurveyllmbased}. Pairwise LLM-as-a-judge has attracted particular attention due to its natural alignment with applications such as automatic model evaluation~\citep{alpaca_eval} and reinforcement learning from AI feedback~\citep{rlaif}. For these applications to be trustworthy, however, the judge model must provide fair and unbiased decisions that align well with human preferences.

Unfortunately, prior research has revealed that LLM judges exhibit multiple types of systematic biases that compromise their reliability~\citep{mt-bench, prejudice-bias, position-bias, shi2025judgingjudgessystematicstudy}. One of them is language bias, where LLMs are demonstrated to show undesired preferences when given texts written in different languages. Specifically, several recent works~\citep{mllm-1, mllm-2, mllm-3} have revealed that in a pointwise judging scenario, LLMs can assign higher scores to answers written in higher-resource languages.

Despite growing awareness of language bias, many critical questions remain unanswered. For example, first, existing studies have primarily examined pointwise judging; the extent of language bias in pairwise comparison settings, where two options are directly compared, remains unclear. Second, it is unknown whether language bias is uniform across domains or more pronounced in subjects tied to cultural and linguistic contexts (such as history or social sciences) compared with language-agnostic domains like mathematics or physics. Third, it is yet unclear whether language bias can be attributed to low-perplexity bias~\citep{self-preference, low-perplexity}, a previously identified bias of LLM-as-a-judge to favor answers with low perplexity. In other words, is language bias simply a special case where options in high-resource languages happen to have lower perplexity, and thus undesirably preferred by the judge model?

In this paper, we present a comprehensive investigation of language bias in pairwise LLM-as-a-judge. We study two distinct scenarios, shown in Figure~\ref{fig:1-pipeline}: same-language judging, where both candidate answers are written in the same language, and inter-language judging, where the two answers are presented in different languages. For same-language judging, we examine performance disparities across languages. For inter-language judging, we investigate whether models exhibit systematic preferences for answers written in high-resource languages. For both settings, we also study the factors that affect the extent of language bias. Finally, we conduct an analysis to determine whether language bias can be reduced to low-perplexity bias.

\section{Related Works}
\subsection{LLM-as-a-Judge}
Most previous works~\citep{li2024llmsasjudgescomprehensivesurveyllmbased} apply LLM-as-a-judge in three different forms: pointwise judging, pairwise judging, and listwise judging. In this paper, we focus on pairwise judging, which is the process where the judge LLM is prompted to choose a better one from two candidate text inputs. Pairwise judging has drawn significant attention recently for its potential applications such as automatic model evaluation~\citep{alpaca_eval, auto-arena}, learning from AI feedback~\citep{rlaif, cgpo}, etc. 

A bottleneck of the application of pairwise LLMs to broader application lies in their performance, often measured by their correlation with human preference. In light of this, some previous works proposed expert models for LLM-as-a-judge, such as Auto-J~\citep{auto-j}, JudgeLM~\citep{judgelm}, Prometheus~\citep{prometheus}, Prometheus-2~\citep{kim2024prometheus2opensource}, and multilingual expert judge model M-Prometheus~\citep{mprometheus}. Other previous works analyze the cause of mismatch with human preferences, and identify certain biases commonly observed in LLM-as-a-judge.

\subsection{Biases of LLM-as-a-Judge}
Previous studies have identified a series of biases in LLM-as-a-judge, which limit the accuracy and potential application scope of the technique. For example, position bias~\citep{position-bias} is a well known issue in pairwise LLM-as-a-judge, where the judge LLM tends to unproportionately favor the option at a certain position. 

One line of previous research studies \textbf{self-preference bias}~\citep{self-preference}, which shows that LLMs tends to unfairly favor options that are generated by themselves. Another previous work~\citep{low-perplexity} show that the favor of the models' own generations can be potentially generalized to the favor of low-perplexity options over high-perplexity ones for the judge LLM. This indicates that judge LLMs might favor the responses that they are more ``familiar'' with rather than the ones that are factually better.

Another type of bias in LLM-as-a-judge is \textbf{language bias}. It has been shown that in a pointwise LLM-as-a-judge setting where the judge model is prompted to assign a score to a given input, the judge model tends to assign lower scores to inputs written in low-resource languages~\citep{mllm-1, mllm-2, mllm-3, mllm-4}. However, a lot of questions remain unclear. For example, (1) whether this phenomenon generalizes to pairwise LLM-as-a-judge, (2) is this phenomenon more significant in specific settings (e.g. in texts related to social or cultural backgrounds compared with texts that are cultural-agnostic such as mathematics or physics), and (3) whether this phenomenon is another special case of the low-perplexity bias described above.

\section{Experiment Settings}
\subsection{Dataset}\label{sec:dataset}

In our experiments, we use the MMMLU dataset~\citep{mmmlu}. MMMLU is a multilingual multiple-choice question answering benchmark obtained by professionally translating the MMLU~\citep{mmlu} test set into 14 languages. Specifically, 14 target languages span multiple regions: Europe (DE, ES, FR, IT, PT), Asia (ZH, JA, KO, HI, BN, ID), the Middle East (AR) and Africa (SW, YO). Consequently, the dataset provides 14,042 parallel questions per language across 57 subjects. More details regarding dataset preprocessing are discussed in Appendix~\ref{app:dataset}.

\subsection{Compared Models}

In our experiments, we study the language bias of a series of LLMs when they are used as judges. The studied models include one closed-source model (GPT-5.1~\citep{openai2025gpt51systemcard}), three general open-source models (Qwen3-30B-A3B, Qwen3-235B-A22B~\citep{qwen3technicalreport}, Llama-3.3-70B-Instruct~\citep{llama3}), one open-source model specifically designed for multilinguality (Aya-expanse-32B~\citep{ayaexpanse}), and two expert LLM-as-a-judge models, including one general judge (Prometheus-8$\times$7B-v2.0~\citep{prometheus}) and one multilingual judge (M-Prometheus-14B~\citep{mprometheus}). More details regarding setup and hyperparameters of the models are in Appendix~\ref{app:prompt-template}.

\begin{figure*}[th!]
    \centering
    \includegraphics[width=\linewidth]{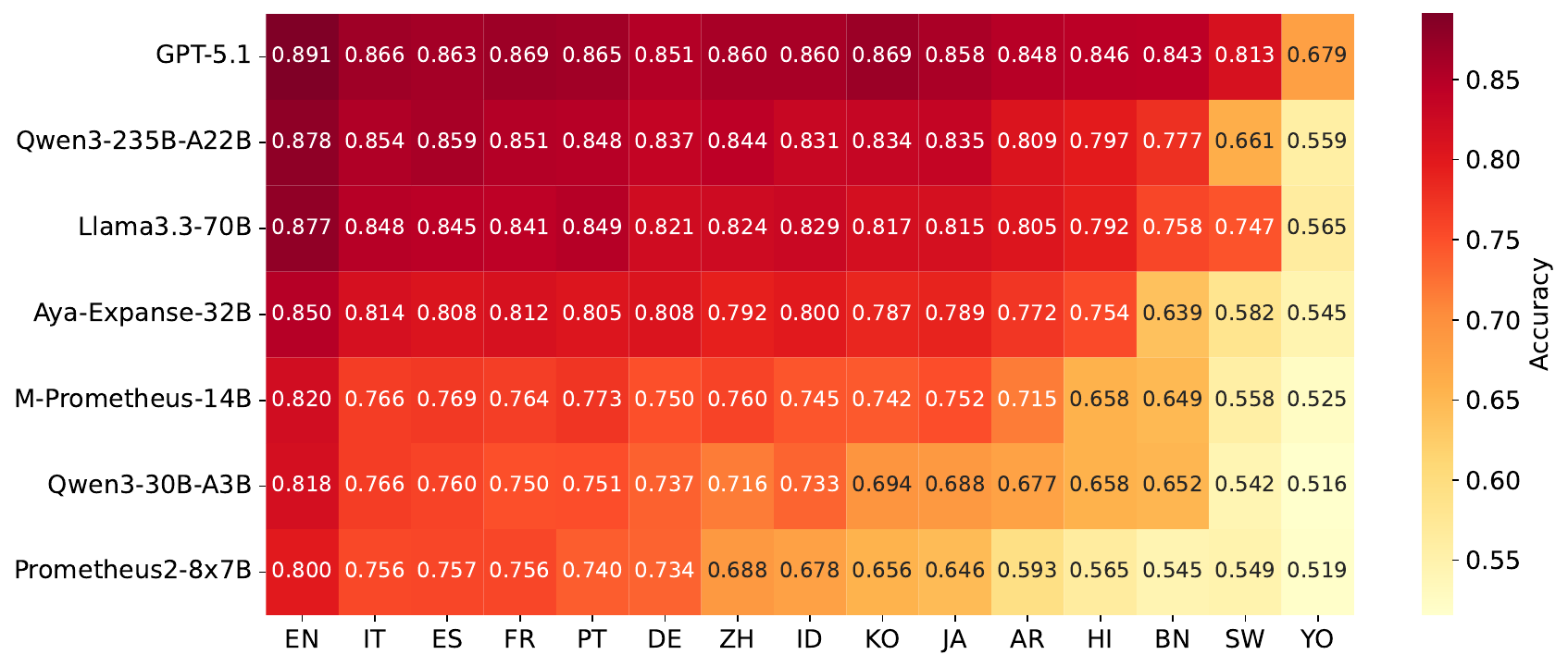}
    \caption{Same-language judging performance of tested models on MMMLU. Models are sorted by average performance across all languages (top to bottom), languages are sorted by average performance across all models (left to right).}
    \label{fig:4-1-model-language}
\end{figure*}

\section{Same-Language Judging: Performance Disparities Between Languages}

In this section, we study how LLM-as-a-judge based on the tested LLMs perform when both answers are presented in the same language. Our goal is to investigate performance disparities across different languages in pairwise LLM-as-a-judge evaluation.

\subsection{Model and Language Performances}\label{sec:same-language-performance}

We first measure the performance of all tested LLMs across all 15 languages in MMMLU. In order to factor out the effect of position bias, we conduct two experiments with swapped positions of the two compared answers, and report the average accuracy of each model. The detailed prompt templates are shown in Appendix~\ref{app:prompt-template}. The results are shown in Figure~\ref{fig:4-1-model-language}. From the results, we can observe the following.

\paragraph{Model performance strongly correlates with region of language.} We observe that the languages with highest average model performances are all European languages (e.g. EN, IT), followed by Asian languages (e.g. ZH, ID), and finally African languages (SW, YO). This holds even for higher-resource Asian languages like Japanese (JP) or Chinese (ZH), compared with lower-resource European languages like Italian (IT)~\citep{commoncrawl-languages-stats}. This indicates that other factors such as proximity to English might be significant in the models' performance in pairwise LLM-as-a-judge, other than the representations of languages in training data.

\paragraph{The relative performance ranking of the models remains consistent across all languages.} For all compared languages, among the open-source models, the model with best performance is always either Qwen3-235B-A22B or Llama3.3-70B, and the model with worst performance is always either Qwen3-30B-A3B or Prometheus2-8x7B. This indicates that models that excel at same-language pairwise LLM-as-a-judge often maintain their relative standing across all languages.

\begin{figure}[th]
    \centering
    \includegraphics[width=\linewidth]{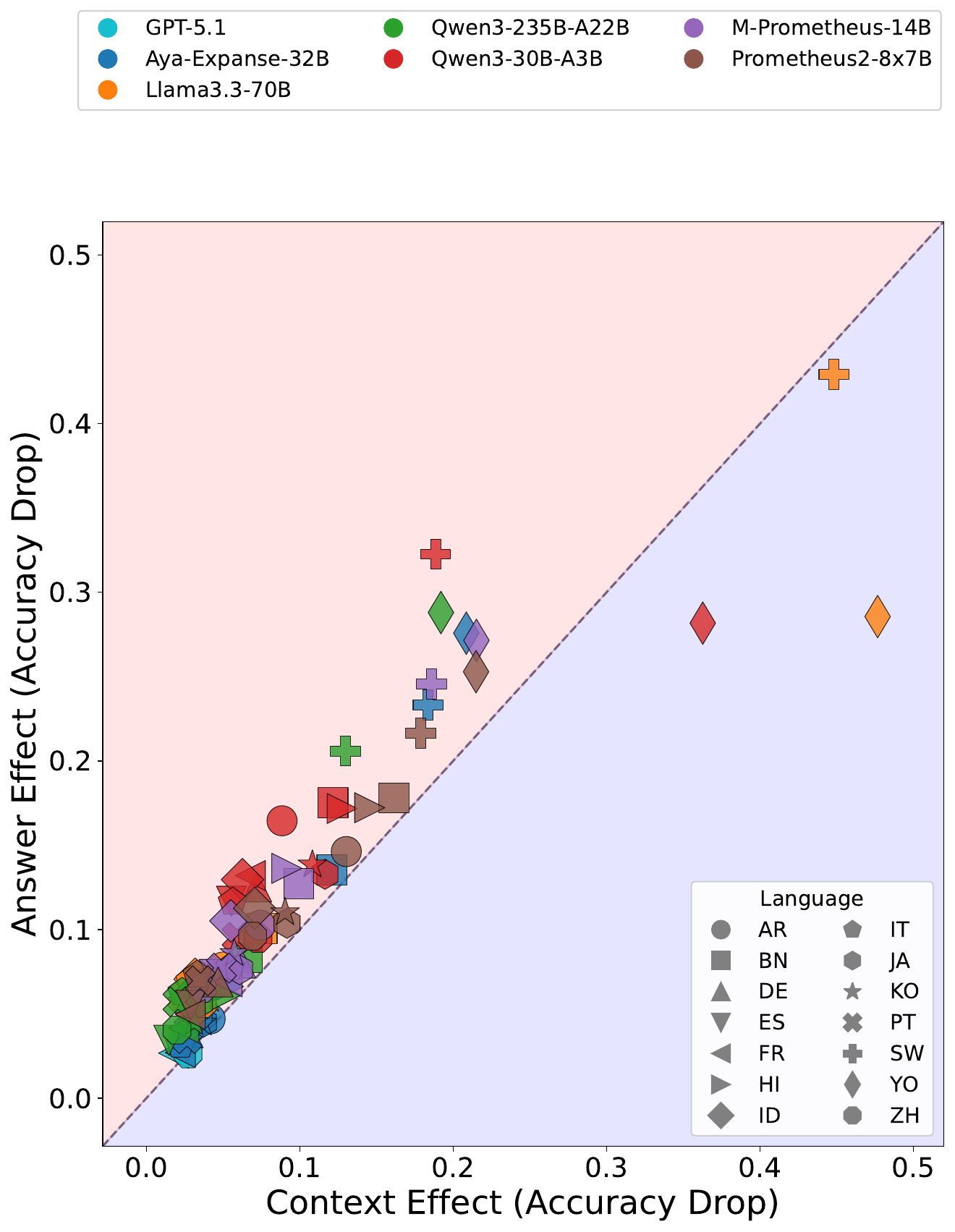}
    \caption{Question Effect versus Answer Effect in same-language judging. Points above the diagonal indicate greater sensitivity to answer language; points below indicate greater sensitivity to question language.}
    \label{fig:4-2-question-answer-effect}
\end{figure}

\subsection{Effect of Question Language and Answer Language}\label{sec:impact-question-answer}
When evaluating multilingual content, language bias can stem from two aspects: the language of the question, or the language of the answer. In this section, we investigate which of these two aspects contribute more to language bias in the same-language comparison setting. Concretely, for each target language in all languages other than English, we conduct experiments with three different settings: English question, English answer; English question, target language answer; target language question, English answer. For each setting, we conduct two sets of experiments with flipped positions and report the average accuracy of each setting. For convenience, we denote the accuracy of the model under these three settings as $\text{Acc}_{EN,EN}$, $\text{Acc}_{EN,TG}$, and $\text{Acc}_{TG,EN}$, respectively. We then calculate the effect of question language and answer language, defined as follows.
\begin{equation}
\text{Question Effect} = \text{Acc}_{EN,EN} - \text{Acc}_{EN,TG}
\end{equation}
\begin{equation}
\text{Answer Effect} = \text{Acc}_{EN,EN} - \text{Acc}_{TG,EN}
\end{equation}

Here, a positive Question Effect shows that translating the question language, from English into other target languages, while keeping the answer in English will reduce the performance. Similarly, a positive score of Answer Effect shows the translated answer while maintaining English Question, which will reduce the performance. The result of each effect reflects the relative contribution of the component to the overall language bias. 

The results are shown in Figure~\ref{fig:4-2-question-answer-effect}. We can see that for most languages and most tested models, the language of the answer is more important than the language of the question. The only exceptions are Llama3.3-70B and Qwen3-30B-A3B on Yoruba (YO) and Swahili (SW), where the models only have near-random performances. 

\begin{figure*}
    \centering
    \includegraphics[width=\linewidth]{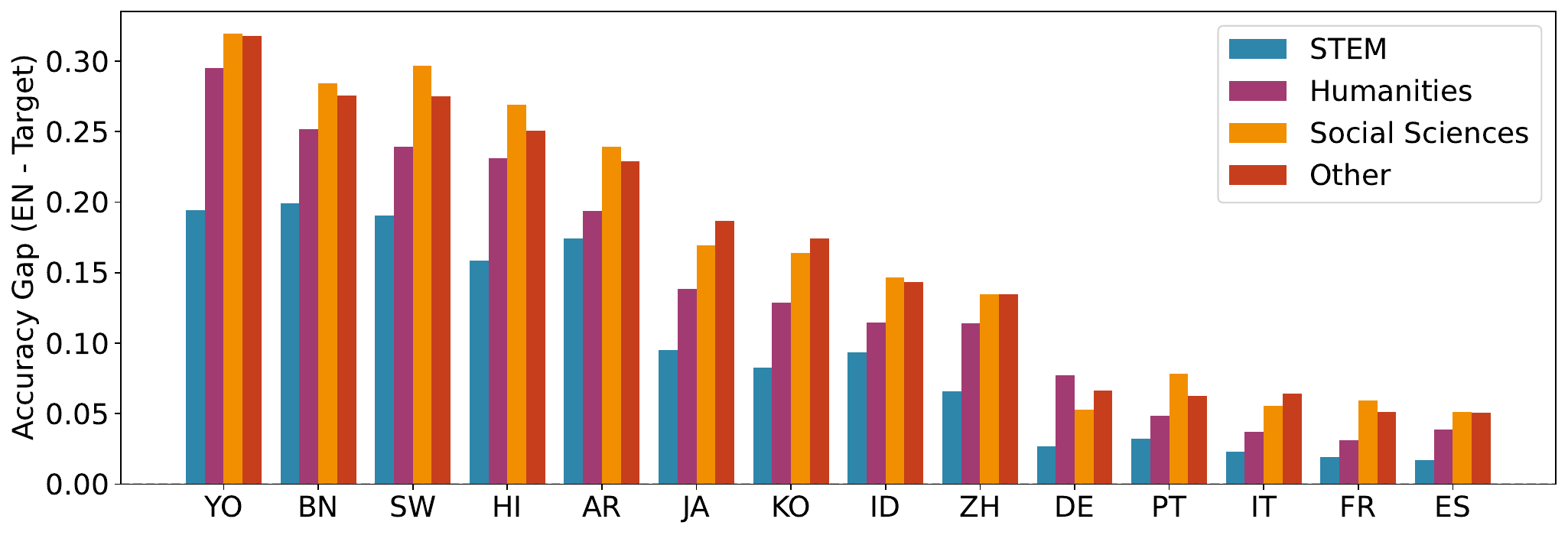}
    \caption{Accuracy gap between English and target language by subject category for Prometheus2-8x7B. Bars represent performance difference across STEM, Humanities, Social Sciences, and Other subjects for each language.}
    \label{fig:4-3-subject-prometheus2}
\end{figure*}

\subsection{Effect of Question Subject}\label{sec:subject}
As we discussed in Section~\ref{sec:dataset}, the MMMLU dataset contains 57 subjects, which can be divided into 17 subcategories. Each subcategory fall into one of the four categories: STEM, humanities, social sciences, and others. In this section, we investigate whether language bias is less significant for subjects that are not related to social or cultural backgrounds (e.g. mathematics) compared with those that are (e.g. culture).

In order to compare language bias in different categories of subjects, we calculate the accuracy of each tested LLM and each language averaged over each subject subcategory. The results for Prometheus2-8x7B in Figure~\ref{fig:4-3-subject-prometheus2} and full results in Appendix~\ref{app:subject} show that language bias is lower for STEM subjects, and higher for Humanities and Social Sciences. As shown in Figure~\ref{fig:4-3-subject-category}, the subjects with highest language bias include Psychology, Business, and Economics, which are all Social Science or Humanities subjects. On the contrary, the subjects with lowest language bias are math and engineering. This indicates LLM-as-a-judge shows less performance gap between languages in questions that contain more language-invariant information and less cultural- or social-related context.

\section{Inter-Language Judging: Bias Across Option Languages}
In this section, we evaluate the performance of LLMs when facing answers in different languages simultaneously, which reflects the real-world scenario that LLM-as-a-judge must evaluate multilingual content.

\subsection{Inter-Language Performance}

\begin{figure*}[th!]
    \centering
    \includegraphics[width=\linewidth]{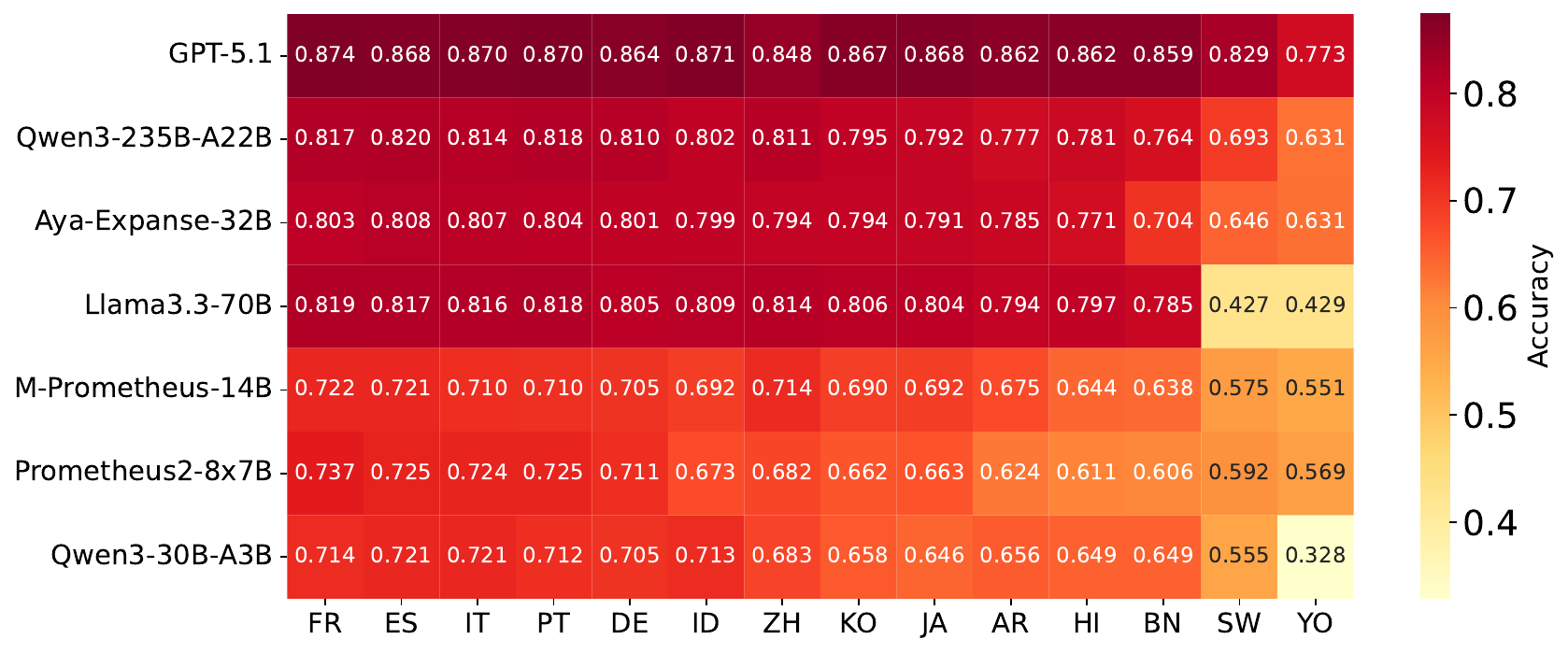}
    \caption{Inter-language judging performance across models and languages. Each cell shows accuracy when comparing English and target language answers, averaged across all four configurations. Models are sorted by average performance across all languages (top to bottom), languages are sorted by average performance across all models (left to right).}
    \label{fig:5-1-inter-language}
\end{figure*}

We first measure the performance of all compared LLMs in the inter-language judging setting. Specifically, in this setting, we consider the case where the model is asked to choose from an English answer and an answer in another language. We perform four sets of experiments as shown in Table~\ref{tab:inter-language-exp}. The prompt templates used are shown in Appendix~\ref{app:prompt-template}. We conduct two experiments for each setting with swapped positions to factor out position bias, and report the average accuracy of the LLM judge. The only exception is GPT-5.1, where we only conduct experiments with the original option order due to budget limit. For simplicity, we denote the results of each setting with $\text{Acc}_{EN,EN\checkmark}$, $\text{Acc}_{EN,TG\checkmark}$, $\text{Acc}_{TG,EN\checkmark}$, $\text{Acc}_{TG,TG\checkmark}$, respectively. Here, $\text{Acc}_{EN,TG\checkmark}$ indicates \textbf{the accuracy on questions written in English where the correct answer is in the target language}. The results are shown in Figure~\ref{fig:5-1-inter-language}. We observe similar conclusions as in the same-language setting reported in Section~\ref{sec:same-language-performance}. First, model performances are strongly correlated with region of language. Second, the relative performance of each model generally persist for all tested languages, though with some exceptions. For example, M-Prometheus-14B has the highest accuracy in English among open-source models, but relatively lower accuracies in other languages. This might be because the Prometheus models are finetuned on same-language comparison scenarios only, and therefore they're less familiar with inter-language settings.

\begin{table}[h]
\centering
\begin{tabular}{lccc}
\toprule
Question & Correct Answer & Incorrect Answer \\
\midrule
English & English & Target \\
English & Target & English \\
Target & English & Target \\
Target & Target & English \\
\bottomrule
\end{tabular}
\caption{Inter-language experiment configurations.}
\label{tab:inter-language-exp}
\end{table}

\subsection{Effect of Question Language and Answer Language}

\begin{figure}[th]
    \centering
    \includegraphics[width=\linewidth]{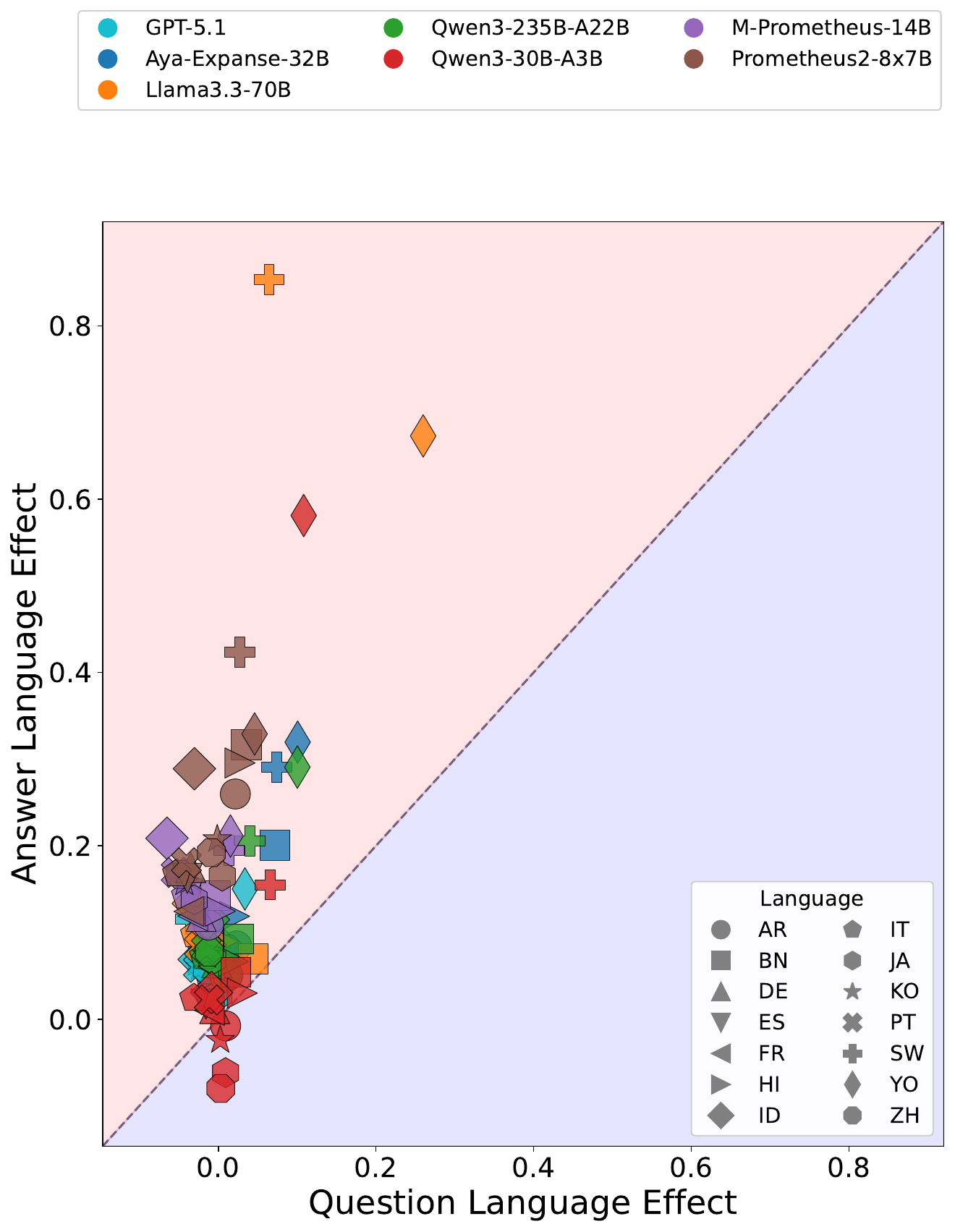}
    \caption{Question language effect versus answer language effect in the inter-language setting. Points above the diagonal indicate more effect caused by answer language; points below indicate more effect caused by question language.}
    \label{fig:5-3-qa-effect}
\end{figure}

In this section, we study the effect of question language and answer language on language bias under the inter-language setting. We use similar definitions and setups as used in Section~\ref{sec:impact-question-answer}. The only difference is that here we consider both when the correct answer is in English or in the target language. Concretely, we calculate the question effect and answer effect as follows:

\begin{align}
\begin{split}
\text{Question Effect} &= \frac{1}{2}[
(\text{Acc}_{EN,EN\checkmark} - \text{Acc}_{TG,EN\checkmark}) \\
&\quad + (\text{Acc}_{EN,TG\checkmark} - \text{Acc}_{TG,TG\checkmark})
]
\end{split} \\
\begin{split}
\text{Answer Effect} &= \frac{1}{2}[
(\text{Acc}_{EN,EN\checkmark} - \text{Acc}_{EN,TG\checkmark}) \\
&\quad + (\text{Acc}_{TG,EN\checkmark} - \text{Acc}_{TG,TG\checkmark})
]
\end{split}
\end{align}

The results are shown in Figure~\ref{fig:5-3-qa-effect}.
From the results, we observe that for languages like Portuguese (PT) or French (FR), the question effect is negative for all the models. This indicates that when conducting inter-language judging involving languages that are comparatively lower-resource but not too low-resource, having the question presented in the target language might help the model perform better. In contrast, the answer effect is mostly positive, which means that the English answers is undesirably preferred, regardless of whether they are truly correct.

\subsection{Answer Language Preference}

From our experiment results, we can also study the following question: is the LLM judge biased towards choosing the English answer compared with answers of other languages? Concretely, we compute the preference of the model towards English answers under two circumstances: when the question is in English ($\text{ENPref}_{EN}$) and when the question is in the target language ($\text{ENPref}_{TG}$).

\begin{align}
    \text{ENPref}_{EN}&=\text{Acc}_{EN,EN\checkmark}-\text{Acc}_{EN,TG\checkmark} \\
    \text{ENPref}_{TG}&=\text{Acc}_{TG,EN\checkmark}-\text{Acc}_{TG,TG\checkmark}
\end{align}

The intuition is that, if the accuracy of the judge model is higher when the correct answers are in English compared with when the answers are in the target language, it means the model is actually inclined to choosing English answers rather than being able to solve every question correctly. We than compare $\text{ENPref}_{EN}$ with $\text{ENPref}_{TG}$ to investigate how does the model's bias change when the question language changes. The results are shown in Figure~\ref{fig:5-4-en-pref}. We can observe that the scatter points mostly fall into the first and fourth quadrants, which shows that in most cases, LLMs are biased towards selecting the English answer when the question is presented in English. Regarding the models' preference when the question is presented in the target language, some models almost always prefer English answers (e.g. Aya-Expanse-32B), some others almost always prefer target language answers (e.g. M-Prometheus-14B), while others fall in the middle where their preferences depend on specific target languages (e.g. Prometheus2-8x7B).

\section{Is Language Bias really Low Perplexity Bias?}

In this section, we examine whether language bias in judging answers stems solely from an LLM's unfamiliarity with a language, as manifested by high perplexity on answers in that language. 

\subsection{Data preprocessing}\label{sec:perp-data-processing}
To investigate this problem, we need to calculate two things. First, we need the judge models' preferences for the first option compared to the other, as defined by the difference in log probability between choosing the two options. Second, we need to compute the judge models' perplexities of each option. This is done through three steps: (1) input the question to the model and get its answer; (2) substitute the answer component in the model's output with the content of the option; (3) calculate the model's perplexity with respect to the option content. We also need to filter the dataset to remove questions that do not fit in this process. Implementation details regarding data preprocessing are shown in Appendix~\ref{sec:perplexity-collection}.

\subsection{Preference-Perplexity Relationship}

We first examine the relationship between LLM judge preference and the perplexity difference between the two options.

\begin{figure}[t]
  \centering
  \includegraphics[width=\columnwidth]{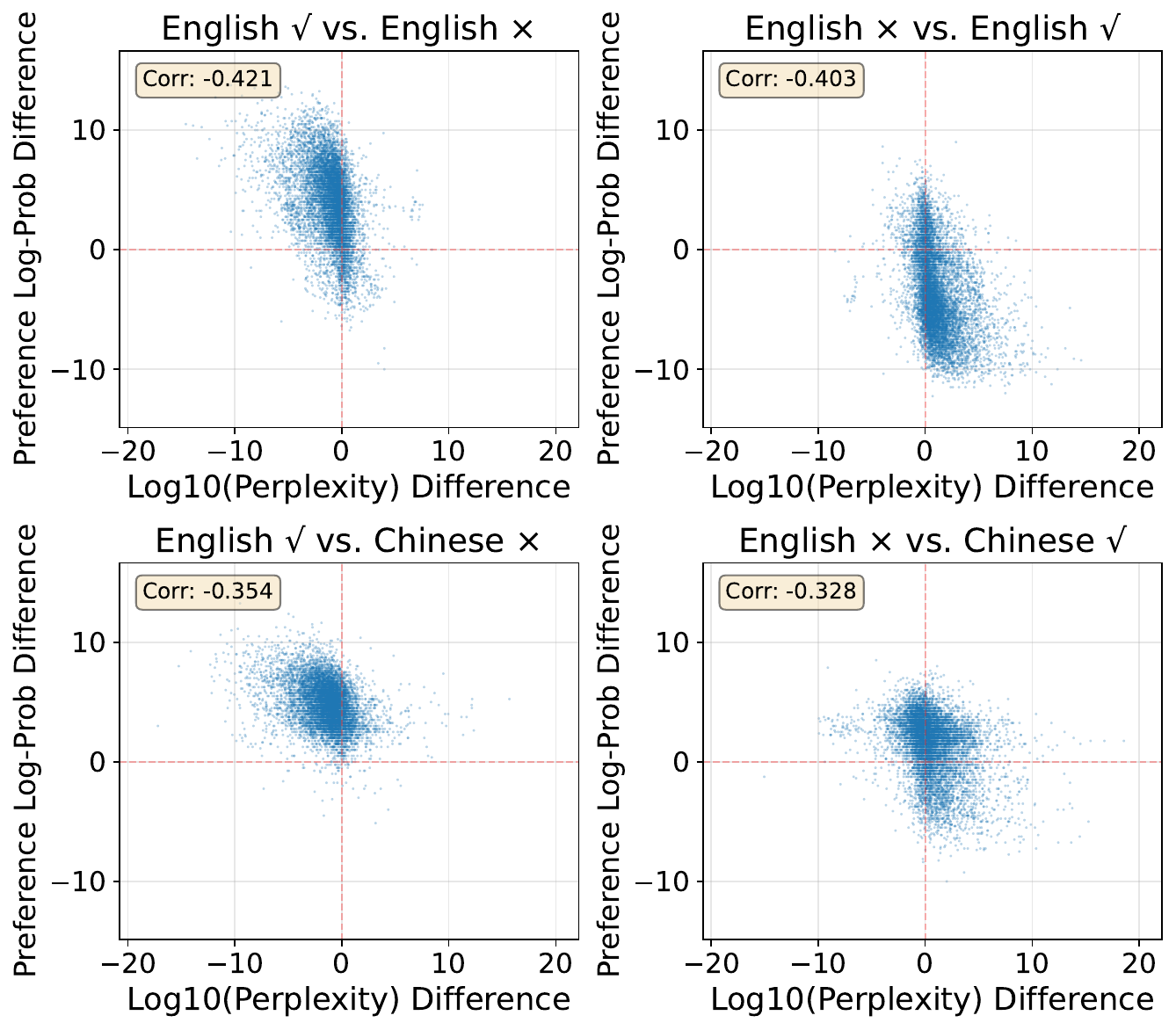}
  \caption{M-Prometheus-14B's preference log probability difference vs. perplexity log difference when the Chinese answer is correct and the English answer is incorrect, with the Pearson correlation coefficient in the yellow box.}
  \label{fig:sample-scatter-plot-figure}
\end{figure}

Figure~\ref{fig:sample-scatter-plot-figure} shows the scatter plot of all data points' model preferences and log perplexity differences across all four inter-language settings described in Table~\ref{tab:inter-language-exp}. We can observe the following:

\paragraph{Slight negative correlation between log perplexity difference and model preference.} For all the settings, we only observe a slight negative correlation of around -0.3 to -0.4 between the two compared variables. This is a initial sign that the preference of the judge model can not be fully explained by perplexity.

\paragraph{Judge model exhibits language bias beyond perplexity.} Comparing the two figures in each column in Figure~\ref{fig:sample-scatter-plot-figure}, we can observe that the scatter plot shifts upward in the top figure compared to the bottom figure in each column. This means that given a fixed log perplexity difference value, the model is more inclined to choosing the first English option when the other answer is in Chinese compared with when the other answer is in English. This further demonstrates that language bias exists even when log perplexity difference between options are the same, which indicates that there exists language bias beyond perplexity.

\subsection{Quantifying Language Bias Beyond Perplexity}

In this section, we quantify how significantly perplexity and other language effects beyond perplexity affect the judging process. We introduce two linear regression models: a reduced model with only perplexity as variable, and a full model with both perplexity and language as variables, represented in Equation~\ref{eq:reduced} and \ref{eq:full}, respectively.

\begin{align}
y &= \alpha_0 + \alpha_1 \,\Delta \text{ppl} \label{eq:reduced} \\
y &= \beta_0 + \beta_1 \,\Delta \text{ppl}
     + \sum_{n=1}^{N} \gamma_n \,\mathbbm{1}(l = n) \label{eq:full}
\end{align}

Here, $y$ denotes signed preference ($y>0$ implies preference for answer 1), $\Delta \text{ppl}$ is the log perplexity difference, and $\mathbbm{1}(l=n)$ indicates whether the current data point is in language $n$ or not, where the value is 1 if yes and 0 if not. We then calculate the \textbf{incremental $R^2$} of the full model compared to the reduced model, which measures how much additional variance of $y$ is explained by the introduction of the language terms $\mathbbm{1}(l=n)$. The results for M-Prometheus-14B are shown in Figure~\ref{fig:prometheus-stacked_bar_tf}, and other results are shown in Appendix~\ref{sec:ftest-results}. In order to verify the statistical significance of the incremental $R^2$, we also conduct a F-test detailed in Appendix~\ref{sec:ftest-details} and \ref{sec:ftest-results}. 

From the results, we observe that low-resource languages such as Yoruba (YO) exhibit the largest incremental $R^2$, indicating that the judge model’s disfavor toward answers in these languages cannot be explained by perplexity alone. In contrast, high-resource languages such as Chinese (ZH) show substantially smaller incremental $R^2$, suggesting that language identity contributes less additional explanatory power beyond perplexity in these cases. Overall, language bias affecting low-resource languages is less attributable to perplexity, whereas for high-resource languages, while language-specific bias beyond perplexity is present, its contribution to the overall variance of model preference is comparatively limited.

\begin{figure}[t]
  \centering
  \includegraphics[width=\columnwidth]{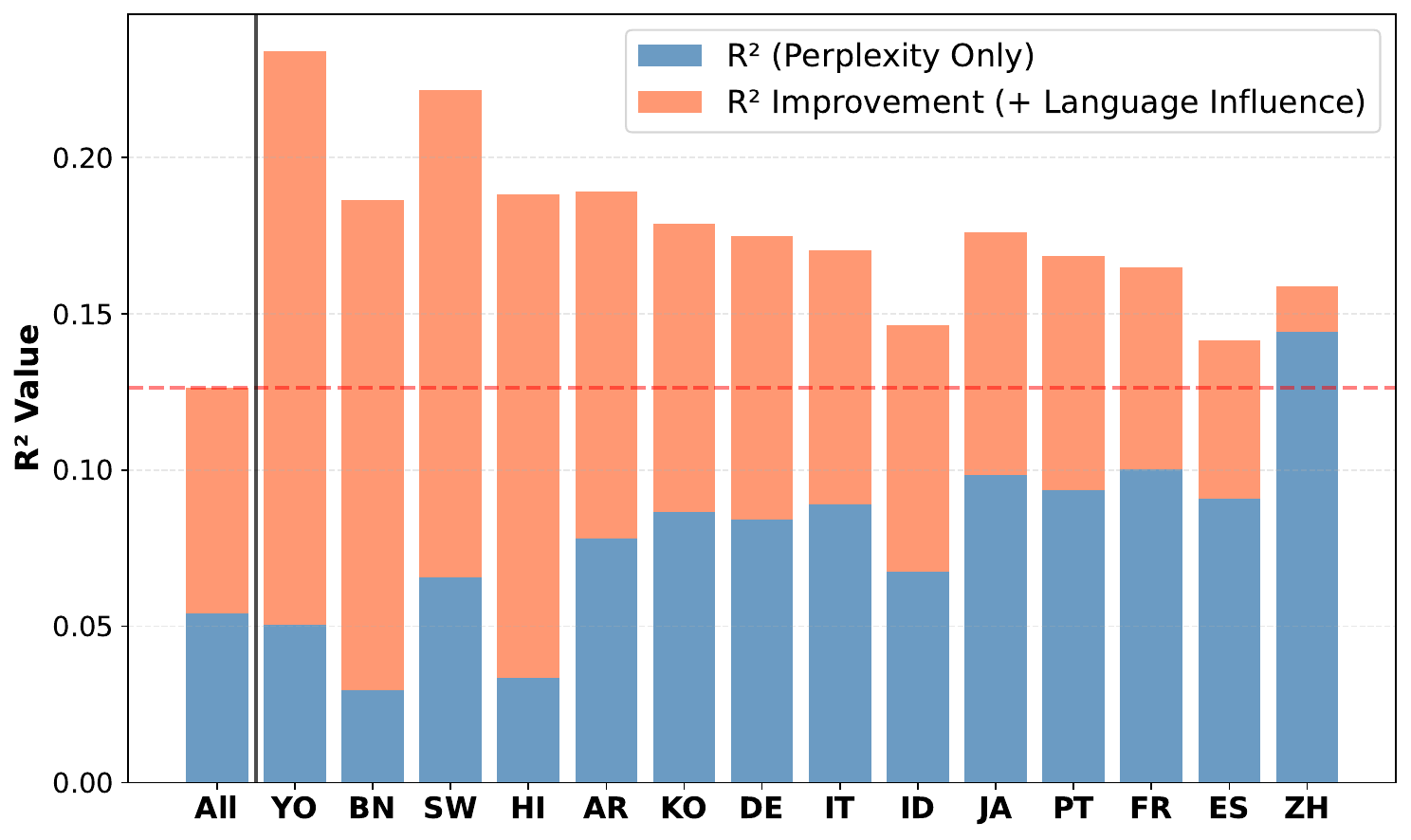}
  \caption{M-Prometheus-14B's $R^2$ decomposition: perplexity contribution (bottom) and additional language contribution (top) when first-language answers are correct and second-language answers are incorrect.}
  \label{fig:prometheus-stacked_bar_tf}
\end{figure}

\section{Conclusion}

In this paper, we conducted a systematic study of language bias in pairwise LLM-as-a-judge. We studies two different settingsL same-language judging and inter-language judging. 

Our analysis reveals several consistent patterns. First, in same-language judging, we observe substantial performance disparities across languages from different regions, with European languages consistently outperforming Asian and African languages. These disparities are more pronounced in culturally and socially grounded subjects that involve more language-specific features than in STEM domains. 

Second, in inter-language judging, most models exhibit a clear preference for English answers when presented alongside answers in other languages, regardless of which answer is correct. We also show that answer language plays a more dominant role than question language in driving this bias. 

Finally, we investigated whether language bias can be reduced to low-perplexity bias. While we find a measurable correlation between preference and perplexity difference, our regression and F-test analyses demonstrate that language identity explains significant additional variance beyond perplexity alone, especially for low-resource languages. This provides strong evidence that language bias is not merely an artifact of fluency or likelihood, but reflects deeper representational and evaluative asymmetries within judge models.

In summary, our findings highlight that language bias remains a fundamental limitation of current pairwise LLM-as-a-judge systems. We hope that our findings provide insights for future research and applications in multilingual LLM-as-a-judge.

\section*{Limitations}

Limitations and potential future directions of our paper include the following:
\begin{itemize}
    \item In the data preprocessing process in Section~\ref{sec:perp-data-processing}, we cannot fully mitigate the portion of perplexity that are not related to the answer choices, like answer framing style mismatch and answer phrasing style mismatch. We also cannot manually verify the correctness of the automatically generated ``styled responses'' by GPT-4.1 because of the infeasible workload. Therefore, there might be noises in the collected perplexity data that may cause the $R^2$ in the reduced linear regression to be smaller than the ground truth value, and causing the F-statistic of the languages to be falsely higher.
    \item Most recent LLMs include reasoning, which could potentially influence the output of LLM-as-a-judge. In our experiments, although we included GPT-5.1 with reasoning mode, we did not consider the potential impact of reasoning in our perplexity experiments.
\end{itemize}

\bibliography{custom}

\appendix

\section{Dataset Preprocessing}\label{app:dataset}
As discussed in Section~\ref{sec:dataset}, in this paper, we conduct pairwise LLM-as-a-judge on MMMLU. In order to convert MMMLU into a pairwise comparison format, we keep the ground-truth (GT) option as the chosen answer, and choose the adjacent incorrect answer as the rejected answer in a cyclic order, for example, if GT is A, the rejected answer is B; if GT is D, then the incorrect answer is A. The total number of questions in the processed dataset remains unchanged.

In our experiments, we require the dataset to contain questions and options that are written in the target language, instead of only non-language-relevant content such as math formulations. To satisfy this requirement, we first preprocess the original MMMLU dataset. Concretely, for each question, we check if all four options contain characters from the target language. Questions will only be retained when they pass the character verification of all 14 target languages simultaneously. After preprocessing, the dataset contain 11,478 parallel questions for each language.

\section{Prompt Templates and Model Configuration}\label{app:prompt-template}
In this section, we list the prompt templates and model configurations used in all our experiments.

\paragraph{Pairwise Comparison}

We adopt a pairwise LLM-as-a-judge paradigm, in which models act as judges to choose a better answer between two compared options. The basic prompt template is:

\begin{promptbox}[Basic Prompt]
\textbf{[System]}\\
You are an impartial judge. Compare the two answers and select the better one.\\
The answer language should not be one of the aspects you consider.\\
Instructions:\\
\hspace*{1em}-- Respond with ONLY the number 1 or 2\\
\hspace*{1em}-- Do NOT write ``Answer 1'' or ``Answer 2''\\
\hspace*{1em}-- Do NOT provide any explanations\\
\hspace*{1em}-- Do NOT write anything else\\[6pt]
\textbf{[User]}\\
Question: \{\textit{question}\}\\
Answer 1: \{\textit{answer\_1}\}\\
Answer 2: \{\textit{answer\_2}\}\\
Your response:
\end{promptbox}

The basic prompt explicitly requires models to ignore answer language, separating the language bias from the fair judgment of the answer.

\paragraph{Inference Configuration}
For all models, we use greedy decoding (temperature = 0.0) with max\_tokens = 3, as the task only requires outputting ``1'' or ``2''. We also collect top-5 log probabilities for further analysis.

\begin{table}[h]
\caption{Experiment setting}
\label{tab:same_lang_4.2}
\centering
\begin{tabular}{lll}
\toprule
\textbf{Experiment} & \textbf{Question Lang} & \textbf{Answer Lang} \\
\midrule
Exp 1 & English & English \\
Exp 2 & Target & English \\
Exp 3 & English & Target \\
\bottomrule
\end{tabular}
\end{table}

\section{Full Results for Effect of Question Subject}\label{app:subject}

\begin{figure*}
    \centering
    \includegraphics[width=\linewidth]{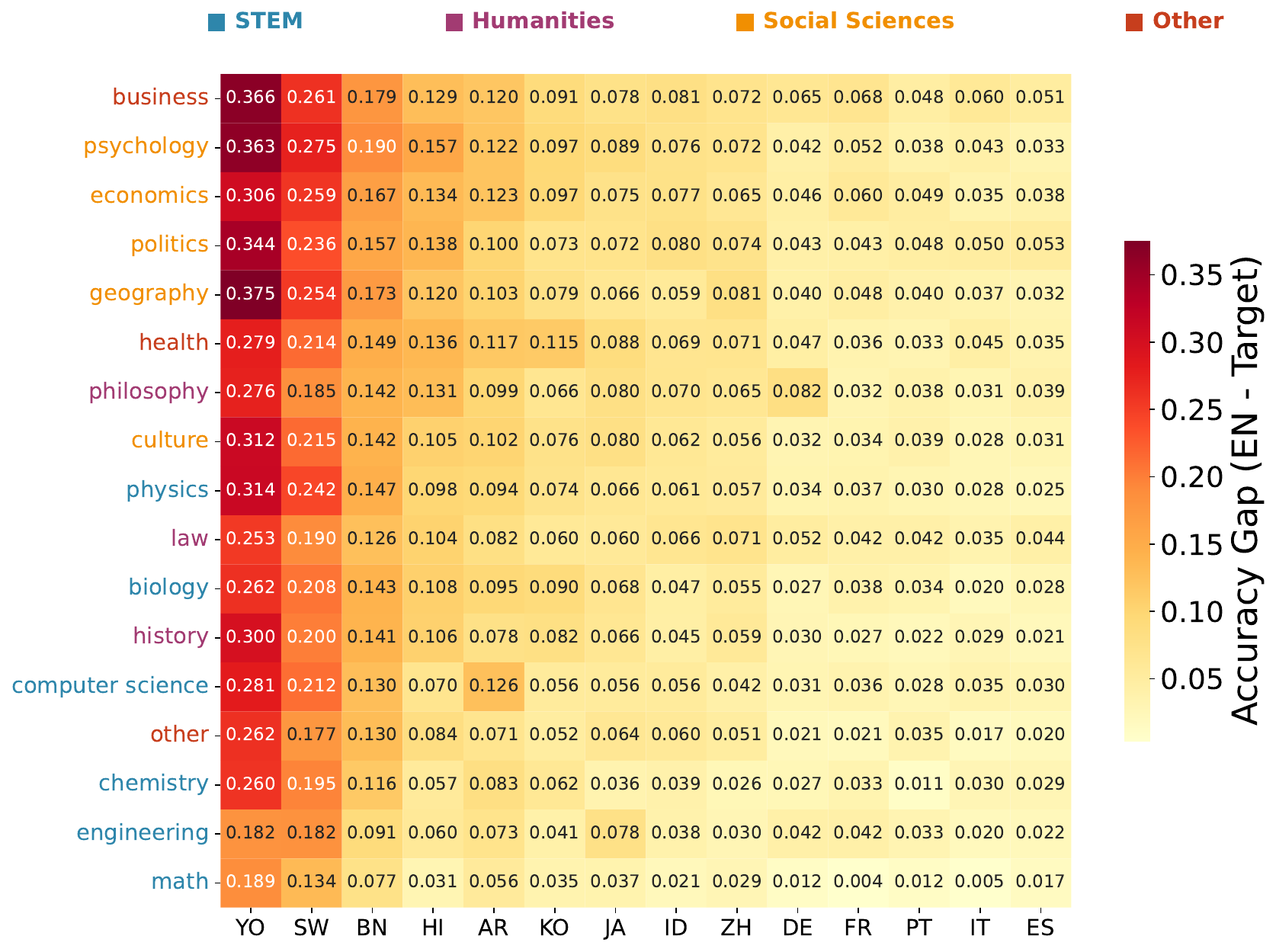}
    \caption{Heatmap of language bias measured as the accuracy gap between English-only evaluation and same-language non-English evaluation. Each cell reports the difference in LLM-as-a-judge performance when both the question and answers are in English versus when both are in the target language, across subject subcategories and languages.}
    \label{fig:4-3-subject-category}
\end{figure*}

\begin{figure*}[t]
    \centering

    \begin{subfigure}{\linewidth}
        \centering
        \includegraphics[width=\linewidth]{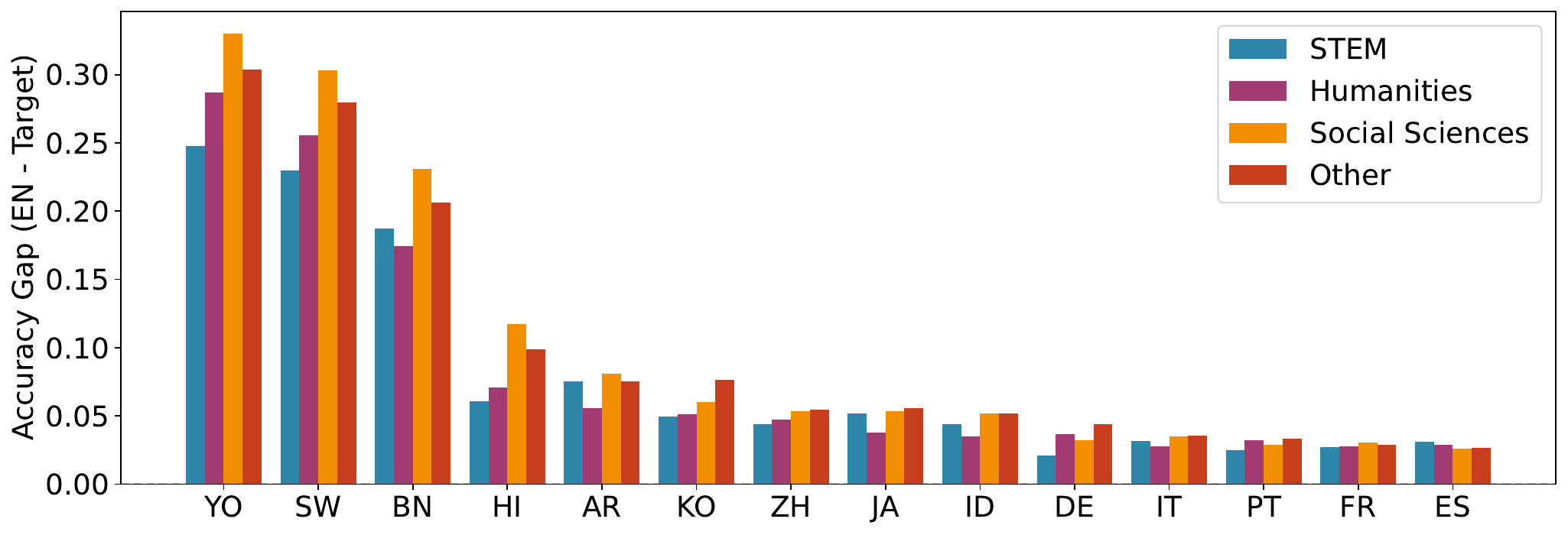}
        \caption{Aya-Expanse-32B}
    \end{subfigure}

    \begin{subfigure}{\linewidth}
        \centering
        \includegraphics[width=\linewidth]{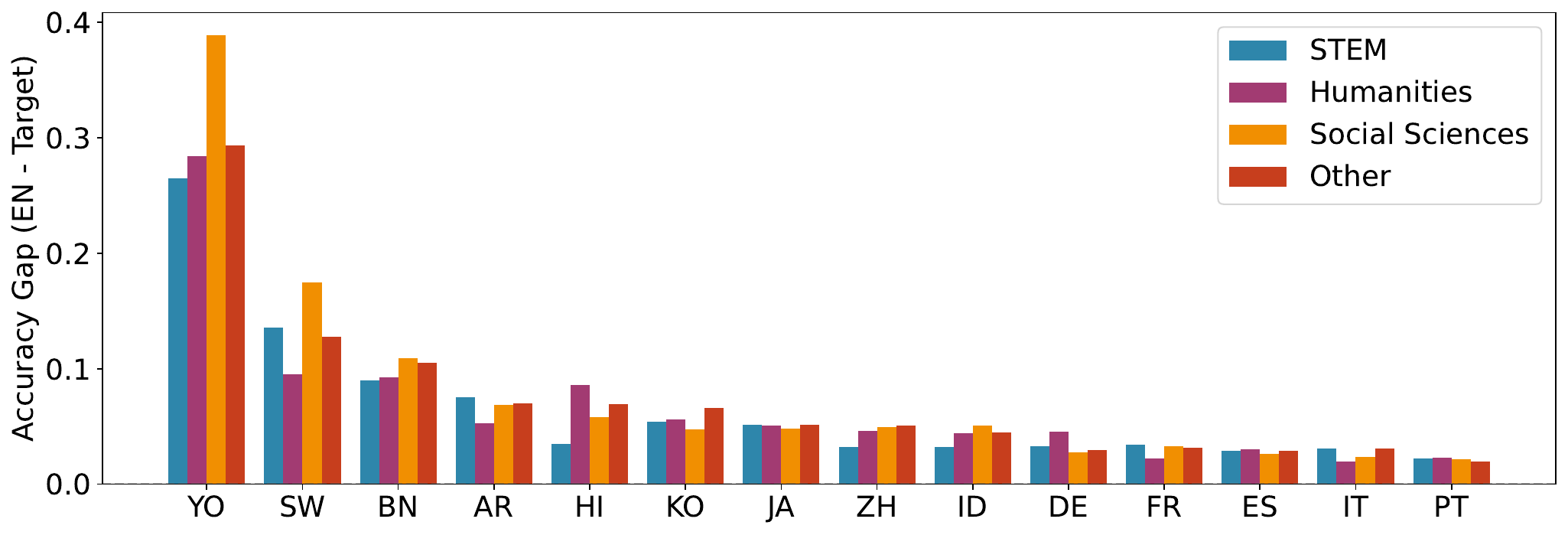}
        \caption{Llama3.3-70B}
    \end{subfigure}

    \begin{subfigure}{\linewidth}
        \centering
        \includegraphics[width=\linewidth]{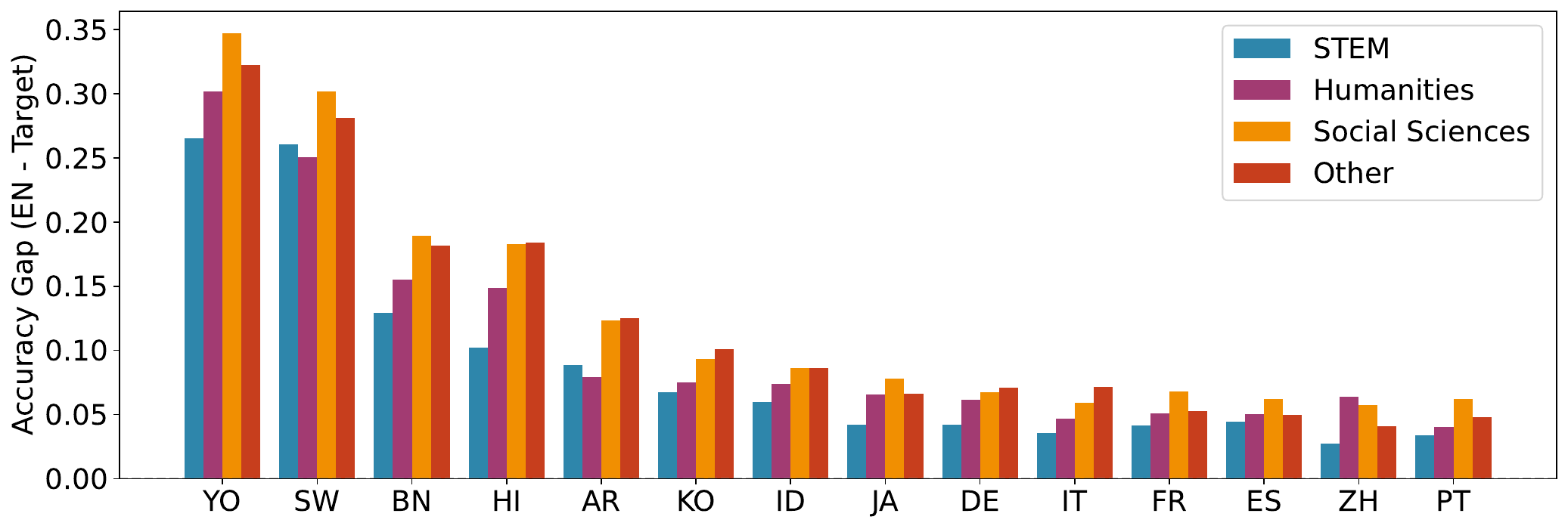}
        \caption{M-Prometheus-14B}
    \end{subfigure}

    \caption{Language bias heatmaps for Aya-Expanse-32B, Llama3.3-70B, and Prometheus-14B across subject categories and languages.}
    \label{fig:language_bias_part1}
\end{figure*}

\begin{figure*}[t]
    \centering

    \begin{subfigure}{\linewidth}
        \centering
        \includegraphics[width=\linewidth]{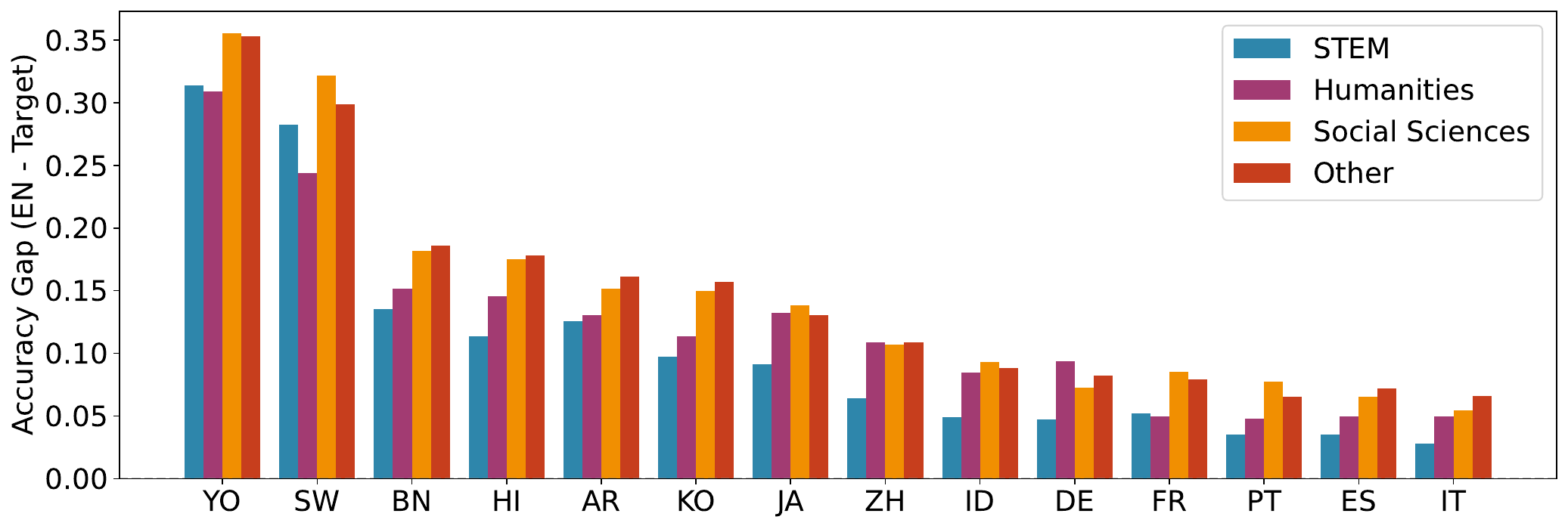}
        \caption{Qwen3-30B-A3B}
    \end{subfigure}

    \begin{subfigure}{\linewidth}
        \centering
        \includegraphics[width=\linewidth]{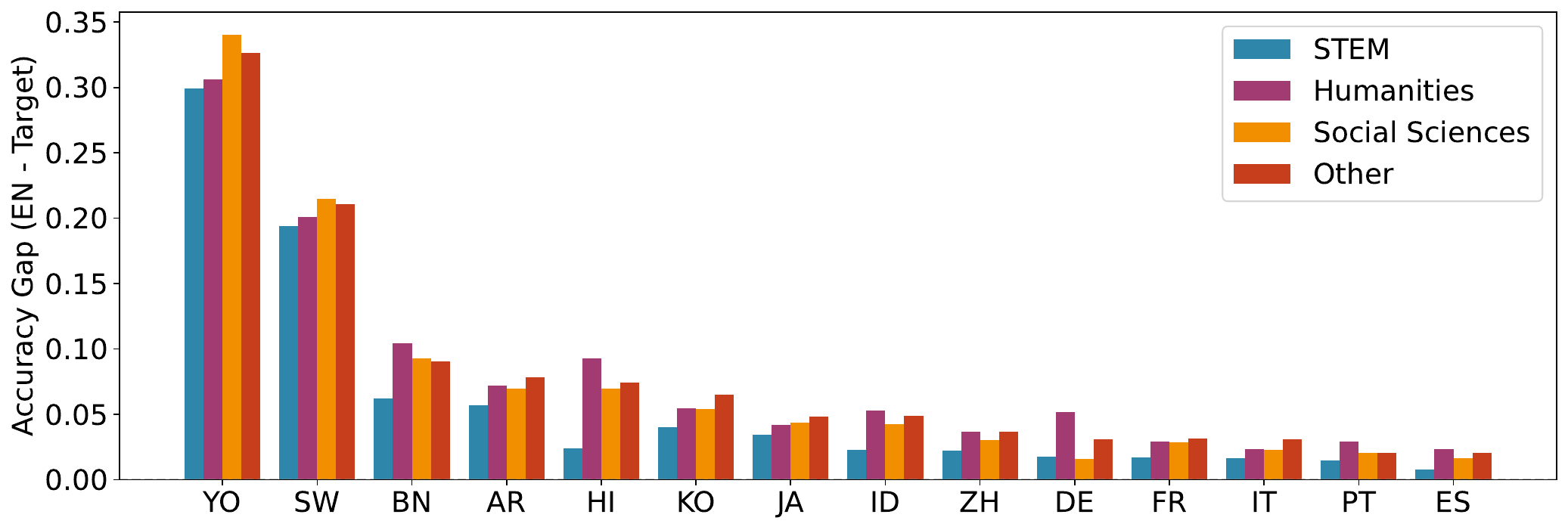}
        \caption{Qwen3-235B-A22B}
    \end{subfigure}

    \caption{Language bias heatmaps for Qwen3 models across subject categories and languages.}
    \label{fig:language_bias_part2}
\end{figure*}

In this section we show the complete result of the question subject experiments in Section~\ref{sec:subject}. The heatmap of language bias averaged over all models for each subject subcategory is shown in Figure~\ref{fig:4-3-subject-category}. The performance of all compared models other than Prometheus2-8x7B (see Figure~\ref{fig:4-3-subject-prometheus2}) on each subject category is shown in Figure~\ref{fig:language_bias_part1} and Figure~\ref{fig:language_bias_part2}.

\section{Full Results for Answer Language Preference}

\begin{figure*}[t]
    \centering
    \begin{subfigure}{0.32\textwidth}
        \centering
        \includegraphics[width=\linewidth]{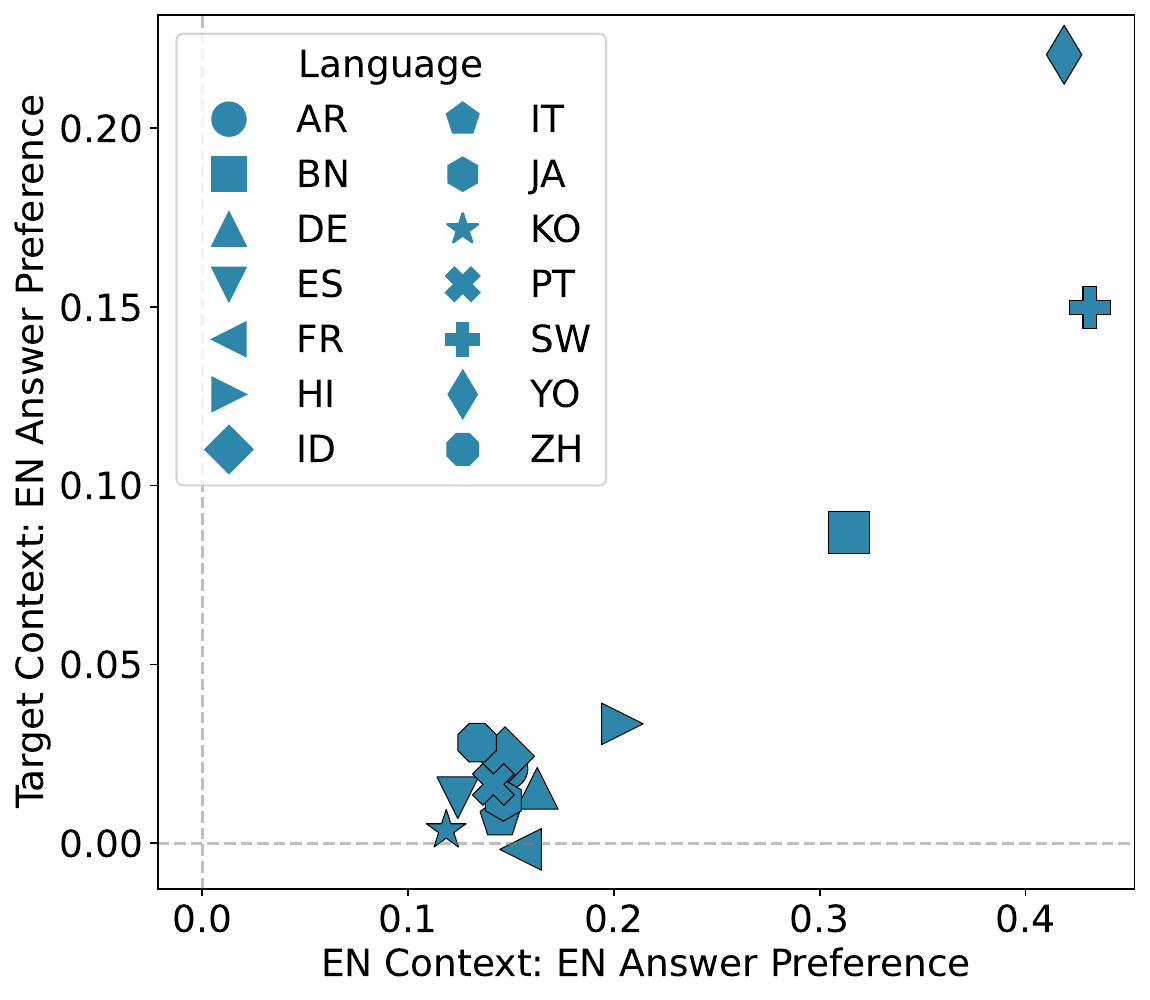}
        \caption{Aya Expanse 32B}
    \end{subfigure}
    \hfill
    \begin{subfigure}{0.32\textwidth}
        \centering
        \includegraphics[width=\linewidth]{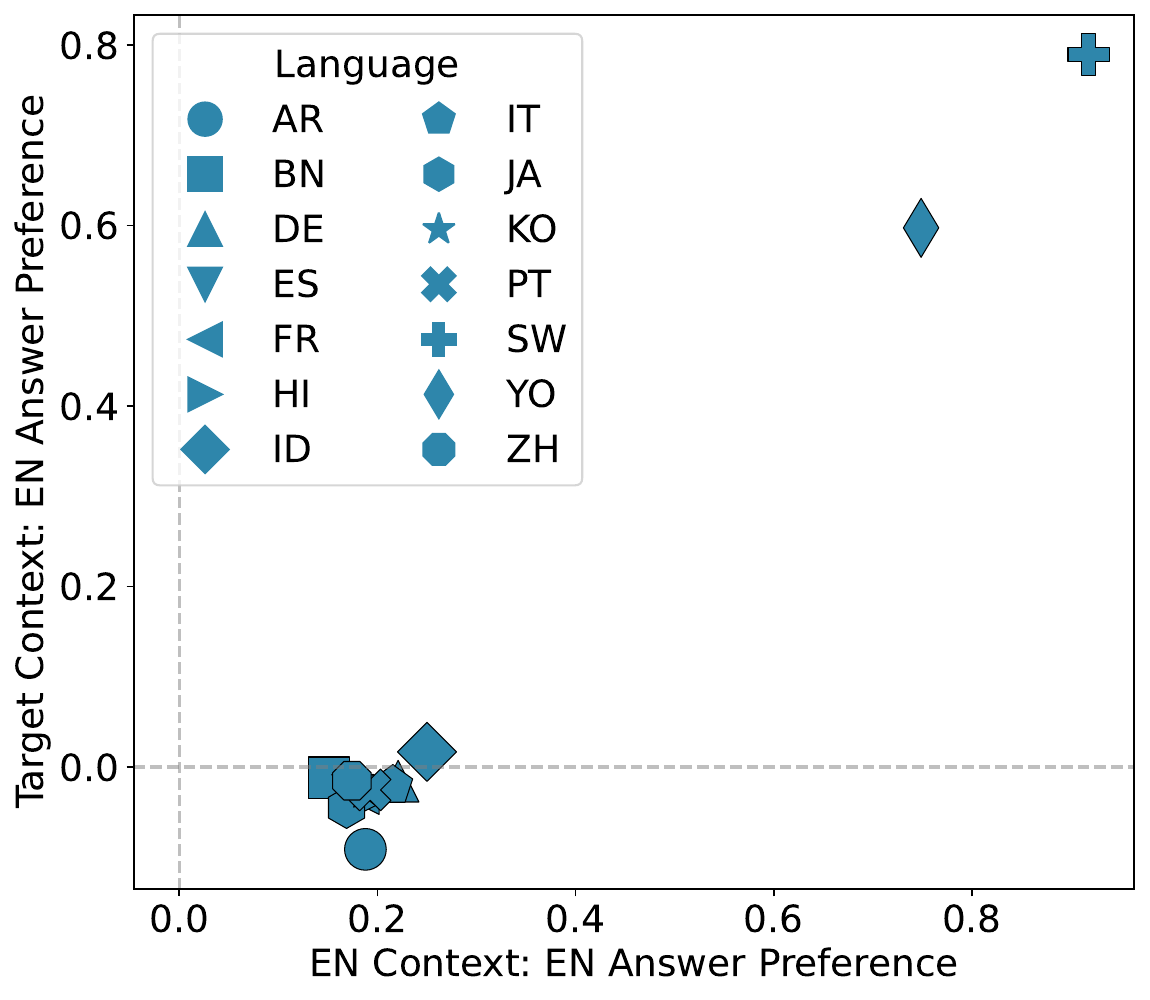}
        \caption{LLaMA 3.3 70B}
    \end{subfigure}
    \hfill
    \begin{subfigure}{0.32\textwidth}
        \centering
        \includegraphics[width=\linewidth]{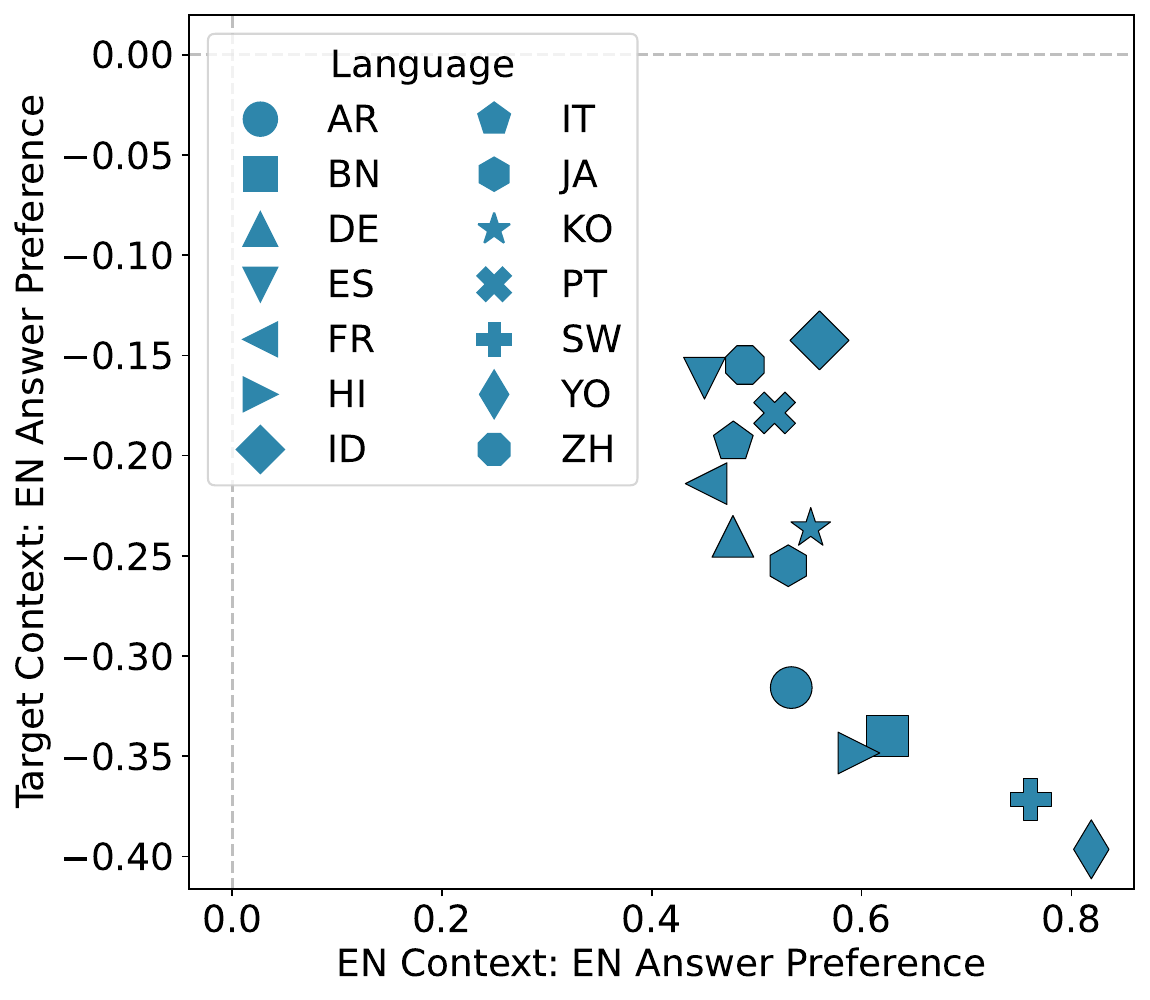}
        \caption{M-Prometheus 14B}
    \end{subfigure}

    \begin{subfigure}{0.32\textwidth}
        \centering
        \includegraphics[width=\linewidth]{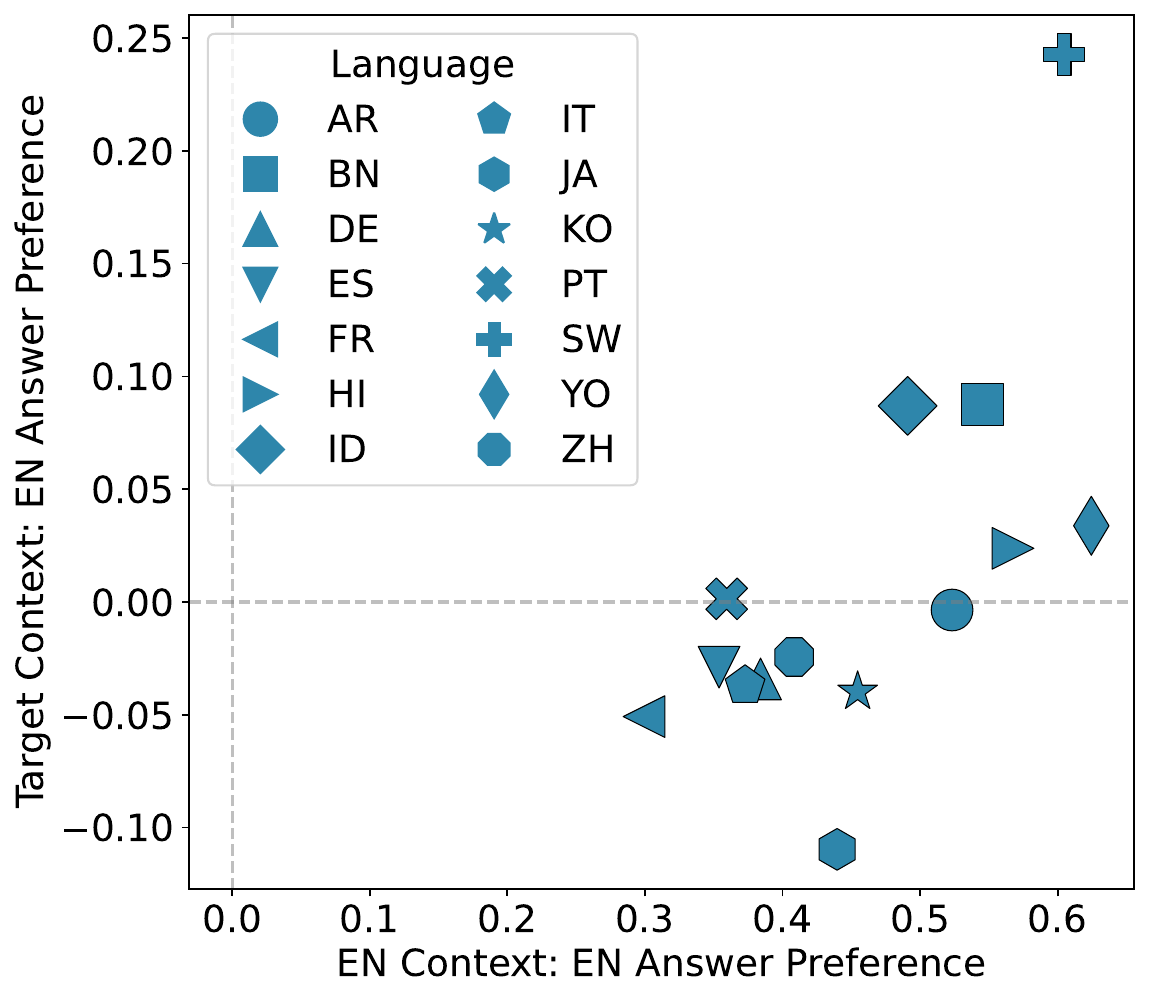}
        \caption{Prometheus2 8×7B}
    \end{subfigure}
    \hfill
    \begin{subfigure}{0.32\textwidth}
        \centering
        \includegraphics[width=\linewidth]{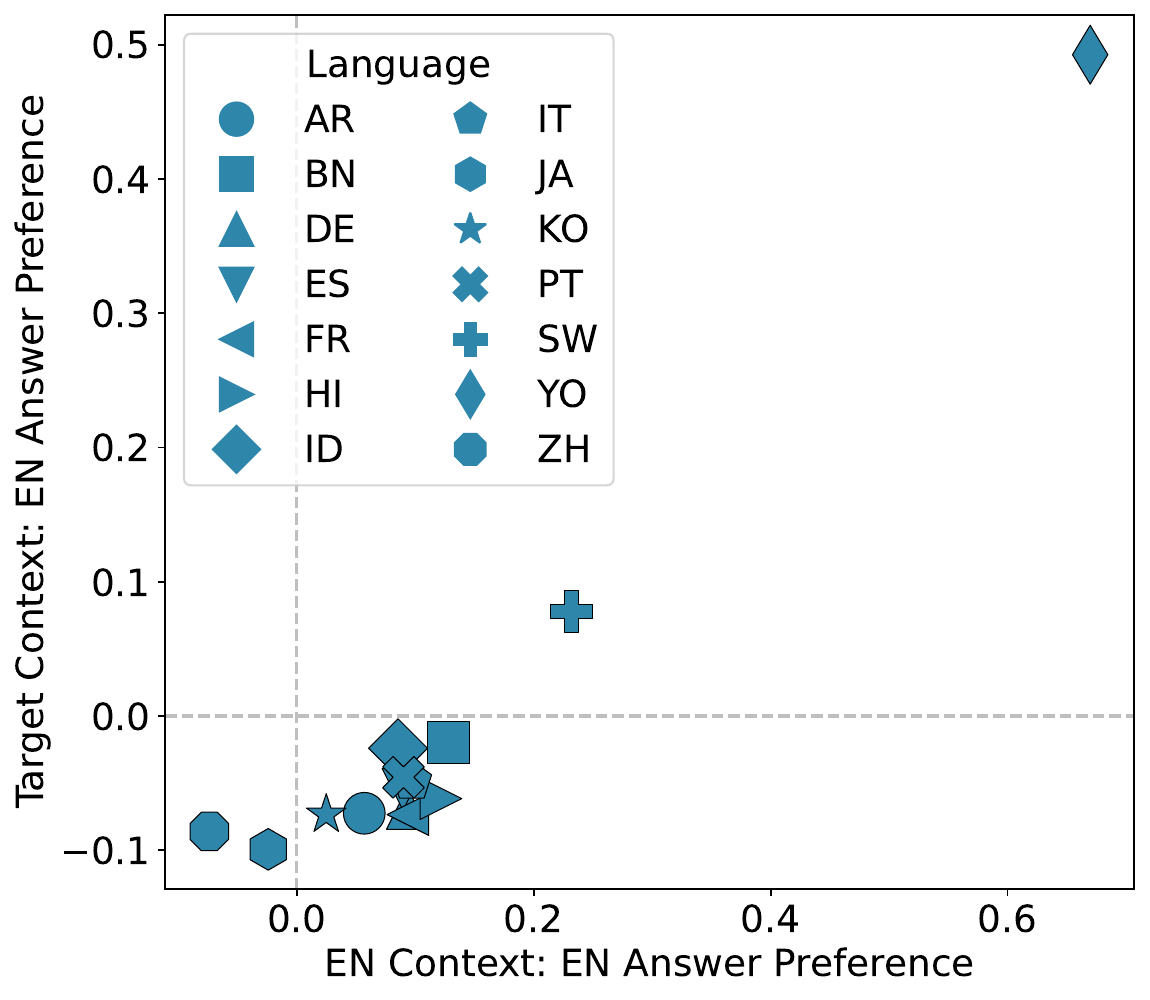}
        \caption{Qwen3 30B A3B}
    \end{subfigure}
    \hfill
    \begin{subfigure}{0.32\textwidth}
        \centering
        \includegraphics[width=\linewidth]{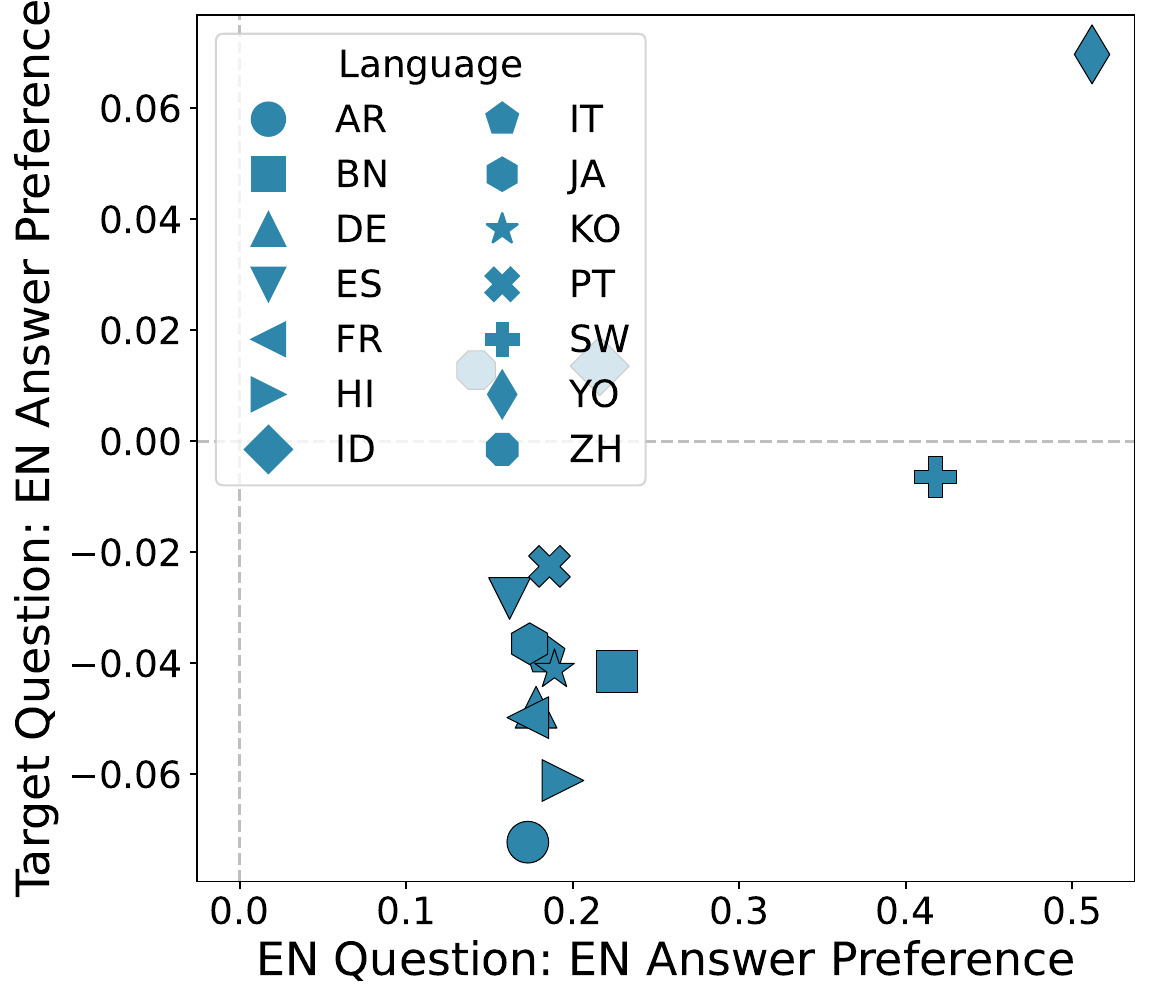}
        \caption{Qwen3 235B A22B}
    \end{subfigure}

    \caption{English answer preferences under different question languages. Each point represents a target language. The x-axis shows $\text{ENPref}_{EN}$, the preference for English answers when the question is in English, and the y-axis shows $\text{ENPref}_{TG}$, the preference for English answers when the question is in the target language.
    }
    \label{fig:5-4-en-pref}
\end{figure*}

In this Section we show the full results of the answer language preference experiment in the inter-language setting. The results are shown in Figure~\ref{fig:5-4-en-pref}.

\section{Perplexity Collection Pipeline Details}
\label{sec:perplexity-collection}

Perplexity measures how ``surprised'' an LLM is by a candidate answer when conditioned on the question, providing a likelihood-based signal of answer compatibility with the model's internal distribution. To isolate the answer content from prompt boilerplate, we compute masked perplexity only over the answer span. Let $x_1, \dots, x_T$ denote the prompt tokens and $m_i \in \{0,1\}$ indicate whether token $i$ belongs to the core answer:
\[
\text{PPL}_{\text{mask}} =
\exp\!\left(
-\frac{1}{\sum_{i=1}^{T} m_i}
\sum_{i=1}^{T} m_i \log p(x_i \mid x_{<i})
\right).
\]

A key challenge is that naively concatenating the question with the dataset answer can yield artificially high perplexity, not because the answer is incorrect, but because it does not match the model's characteristic response style. To address this, we employ a three-step pipeline that fuses the model's natural answering style with the dataset answer:

\begin{enumerate}
\item \textbf{Collect natural response:} Present the question to the LLM without revealing candidate answers to obtain its natural response.
\item \textbf{Fuse response style:} Use GPT-4.1 to synthesize a response that preserves the model's style while replacing substantive content with the dataset answer.
\item \textbf{Generate perplexity:} Feed the synthesized response to the LLM, extract logits over the core answer tokens (marked by ``<answer>'' tags), and compute masked perplexity.
\end{enumerate}

This approach ensures that perplexity reflects the model's assessment of answer content rather than stylistic mismatch. Below we provide implementation details for each step.

\subsection{Step 1: Collecting Natural Responses}

We present each question to the LLM without revealing candidate answers to obtain the model's natural response. These responses typically begin with boilerplate phrasing reflecting the model's stylistic tendencies (e.g., ``The answer is...'' or ``This phenomenon occurs because...'').

\textbf{Prompt used to collect natural responses:}

\begin{promptbox}[Prompt for Generating Natural Responses to Questions]
\textbf{[System]}

Please CONCISELY answer the question in [language\_name] WITHOUT reasoning or explanation.

\textbf{[User]}

[question]
\end{promptbox}

\subsection{Step 2: Fusing Response Style with Dataset Answers}

Given the model's natural response, we use GPT-4.1 to synthesize a response that preserves the model's style while replacing the substantive content with the dataset answer.

\textbf{Prompts used for style fusion:}

\begin{promptbox}[Prompt for Generating Synthesized Responses that Fuses Dataset Answers with an LLM's Response Style]
\textbf{[System]}
You are a helpful assistant. The user is going to provide you a question, an LLM's response to the question, and two extra answers. 

Your task is to merge the style of the LLM's response with the two given answers, and produce two responses that have the same meaning as the given answers but in the style of the LLM's response. 

The two synthesized responses must be IDENTICAL except for the very essence of the answers. 

The essence of the answers in the two responses should be enclosed in <answer> and </answer> tags. 

If the LLM's response contains any content that is related to the decision of the answer, you should discard it in the synthesized responses.

Here is an example:

Question: Judge the following statements: 1+1=3. All integers are either even or odd.

LLM's Response: The first statement is incorrect. 1+1=2. The second statement is **correct**. All integers are either even or odd.

Answer 1: True, True
Answer 2: False, False
Your final output:
{

  "response\_1": "The first statement is <answer>true</answer>. The second statement is **<answer>true</answer>**."
  
  "response\_2": "The first statement is <answer>false</answer>. The second statement is **<answer>false</answer>**."
  
}

Begin your response with your first trial of generating the two synthesized answers WITHOUT using JSON format. Then check the following:

1. Is the essence of the answers enclosed in <answer> and </answer> tags while the rest of the content is in the style of the LLM's response?

2. Apart from the content inside the <answer> tags, are the two responses identical?

3. Does the content outside the <answer> tags reveal the decision of the answers? If so, it should be removed.

4. Are the details from the LLM's response faithfully preserved, including letter cases and special decorations like ``"**" for bold?

5. Did you make up styles or sentence framing that do not exist in the LLM's response? If the LLM's response contains only the direct answer without any framing, then wrap the answer with <answer> tags directly without adding any other styles or sentence framing.

Finally, output the final version of the two responses in JSON format with keys "response\_1" and "response\_2".

\textbf{[User]}

Question: [question]

LLM's Response: [response]

Answer 1: [answer\_correct]

Answer 2: [answer\_incorrect]
\end{promptbox}

\subsection{Example of Styled Response Synthesis}

The following real example illustrates how we synthesize styled responses:

\begin{promptbox}[Example of Styled Response Synthesis]
\textbf{[Question]}

To say that an action is intrinsically permissible (a feature mentioned in the doctrine of double effect) is to say that

\textbf{[LLM's Natural Response]}

the action is morally acceptable in itself, regardless of its consequences.

\textbf{[Correct Answer from Dataset]}

the action, apart from its effects, is morally permissible.

\textbf{[Incorrect Answer from Dataset]}

the action, only because of its effects, is morally permissible.

\textbf{[Synthesized Correct Response]}

the action is <answer>apart from its effects, morally permissible</answer>, regardless of its consequences.

\textbf{[Synthesized Incorrect Response]}

the action is <answer>only because of its effects, morally permissible</answer>, regardless of its consequences.
\end{promptbox}

\subsection{Step 3: Computing Masked Perplexity}

After obtaining synthesized responses, we concatenate the question prompt with each response (removing the ``<answer>'' tags) and feed the combined prompt to the LLM to obtain logits. We construct a mask over the core answer span (previously enclosed by the tags) and compute masked perplexity using only those logits.

\section{F-Test Details}
\label{sec:ftest-details}

To determine whether the language bias observed in judge LLMs can be fully attributed to the model's unfamiliarity with specific languages, we employ a nested F-test framework. Our approach compares two linear regression models: a reduced model that depends solely on perplexity, and a full model that incorporates both perplexity and language-specific effects. By comparing the goodness-of-fit between these models, we can quantify the extent to which language identity contributes to preference bias beyond what perplexity alone can explain.

\subsection{Model Specifications}

We define two linear models:

\textbf{Reduced Model (Perplexity Only):}
\[
y_i = \alpha_0 + \alpha_1 \Delta \text{ppl}_i + \epsilon_i
\]

\textbf{Full Model (Perplexity + Language):}
\[
y_i = \beta_0 + \beta_1 \Delta \text{ppl}_i + \sum_{n=1}^{N} \gamma_n \,\mathbbm{1}(l_i=n) + \epsilon_i
\]

where:
\begin{itemize}
\item $y_i$ is the preference for answer 1 over answer 2, measured as the log probability difference
\item $\Delta \text{ppl}_i$ is the log perplexity difference between the two answers
\item $\mathbbm{1}(l_i=n)$ is an indicator variable that equals 1 if sample $i$ belongs to language pair $n$, and 0 otherwise
\item $N$ is the total number of distinct language pairs in the experiment
\item $\epsilon_i$ is the error term
\end{itemize}

\textbf{Language Pair Encoding:} We use language pairs rather than individual languages because preference statistics inherently require comparing two answers in potentially different languages. Language-specific bias is characterized by how the preference-perplexity relationship varies across language pairs. For instance, if this relationship differs between ``English (correct) vs. English (incorrect)'' and ``English (correct) vs. Chinese (incorrect),'' we can attribute the variance to Chinese-specific effects beyond perplexity.

We use English as the reference language in all pairs because it is the most prevalent language in LLM pre-training corpora. Each indicator variable $\mathbbm{1}(l_i=n)$ corresponds to a language pair ``English vs. language $n$,'' where language $n$ represents one of the 14 non-English languages in our dataset.

The reduced model has $k_{\text{red}} = 2$ parameters ($\alpha_0$, $\alpha_1$), while the full model has $k_{\text{full}} = 2 + N$ parameters ($\beta_0$, $\beta_1$, and $N$ language coefficients $\gamma_n$).

\subsection{Experimental Design}

For each judge model and correctness configuration (e.g., English correct vs. non-English incorrect, or vice versa), we conduct two complementary experiments:

\paragraph{Aggregated Analysis}
We pool samples from all language pairs (English vs. each of 14 non-English languages) into a single aggregated dataset. Each sample retains its language pair identity through indicator variables. The full model learns 16 parameters: one intercept ($\beta_0$), one perplexity coefficient ($\beta_1$), and 14 language pair coefficients ($\gamma_1, \ldots, \gamma_{14}$). This yields a single $R^2$ improvement metric that captures the overall contribution of language identity across all non-English languages.

\paragraph{Pairwise Analysis}
To isolate the effect of each individual language, we perform targeted comparisons between the baseline pair (English vs. English) and each target pair (English vs. language $n$). This setup uses 4 parameters: one intercept, one perplexity coefficient, and two language pair indicators. By measuring how the preference-perplexity relationship shifts from the baseline to each target language, we obtain language-specific $R^2$ improvements for all 14 non-English languages.

\subsection{Model Training}

Both the reduced and full models are trained on identical datasets using the Adam optimizer with the following configuration:
\begin{itemize}
\item Learning rate: 0.001
\item Loss function: mean squared error (MSE)
\item Convergence criterion: early stopping triggered when validation loss plateaus
\end{itemize}

After training, we evaluate model fit using the coefficient of determination:
\[
R^2 = 1 - \frac{\sum_i (y_i - \hat{y}_i)^2}{\sum_i (y_i - \bar{y})^2}
\]
where $\hat{y}_i$ denotes the model's prediction for sample $i$ and $\bar{y}$ is the mean of all observed values. The $R^2$ metric quantifies the proportion of variance in preference scores explained by the model.

\subsection{Statistical Testing}

To assess whether language identity contributes significantly to preference beyond perplexity, we perform a nested F-test comparing the reduced and full models.

\textbf{Hypotheses:}
\begin{align*}
H_0&:\ \gamma_1=\gamma_2=\cdots=\gamma_N = 0 \quad  \\
H_1&:\ \exists\, n \in \{1,\ldots,N\} \text{ such that } \gamma_n \neq 0 \quad 
\end{align*}
where $H_0$ is the hypothesis that language has no additional effect and $H_1$ is the hypothesis that language contributes beyond perplexity.

\textbf{Test Statistic:}
\[
F =
\frac{(R^2_{\text{full}} - R^2_{\text{red}})/(k_{\text{full}}-k_{\text{red}})}
{(1 - R^2_{\text{full}})/(N_{\text{samples}} - k_{\text{full}})}
\]

The numerator captures the incremental variance explained per additional parameter in the full model, while the denominator normalizes by the unexplained variance per degree of freedom. Here, $N_{\text{samples}}$ denotes the total number of samples in the dataset.

Under $H_0$, the F-statistic follows an $F$-distribution with degrees of freedom $(k_{\text{full}}-k_{\text{red}},\, N_{\text{samples}}-k_{\text{full}})$, which simplifies to $F(N,\, N_{\text{samples}}-2-N)$ in our experimental setup.

\subsection{Interpretation}

A statistically significant F-test (large F-statistic, small p-value) rejects $H_0$ and provides evidence that:
\begin{enumerate}
\item \textbf{Language-specific bias exists:} Language identity contributes to preference beyond what perplexity alone can explain.
\item \textbf{Systematic variation across languages:} The model's preference behavior differs systematically across language pairs, even after controlling for perplexity differences.
\item \textbf{Bias is not reducible to familiarity:} The observed language bias cannot be fully attributed to varying perplexity levels (i.e., model unfamiliarity with certain languages).
\end{enumerate}

The magnitude and sign of individual language coefficients $\gamma_n$ reveal language-specific bias patterns. Positive $\gamma_n$ values indicate that the model favors language pair $n$ beyond what perplexity predicts, while negative values indicate disfavor. The $R^2$ improvement quantifies the practical significance of language effects: larger improvements indicate that language identity plays a substantial role in shaping model preferences.

\section{F-Test Results}
\label{sec:ftest-results}

\begin{figure}[t]
  \centering
  \includegraphics[width=\columnwidth]{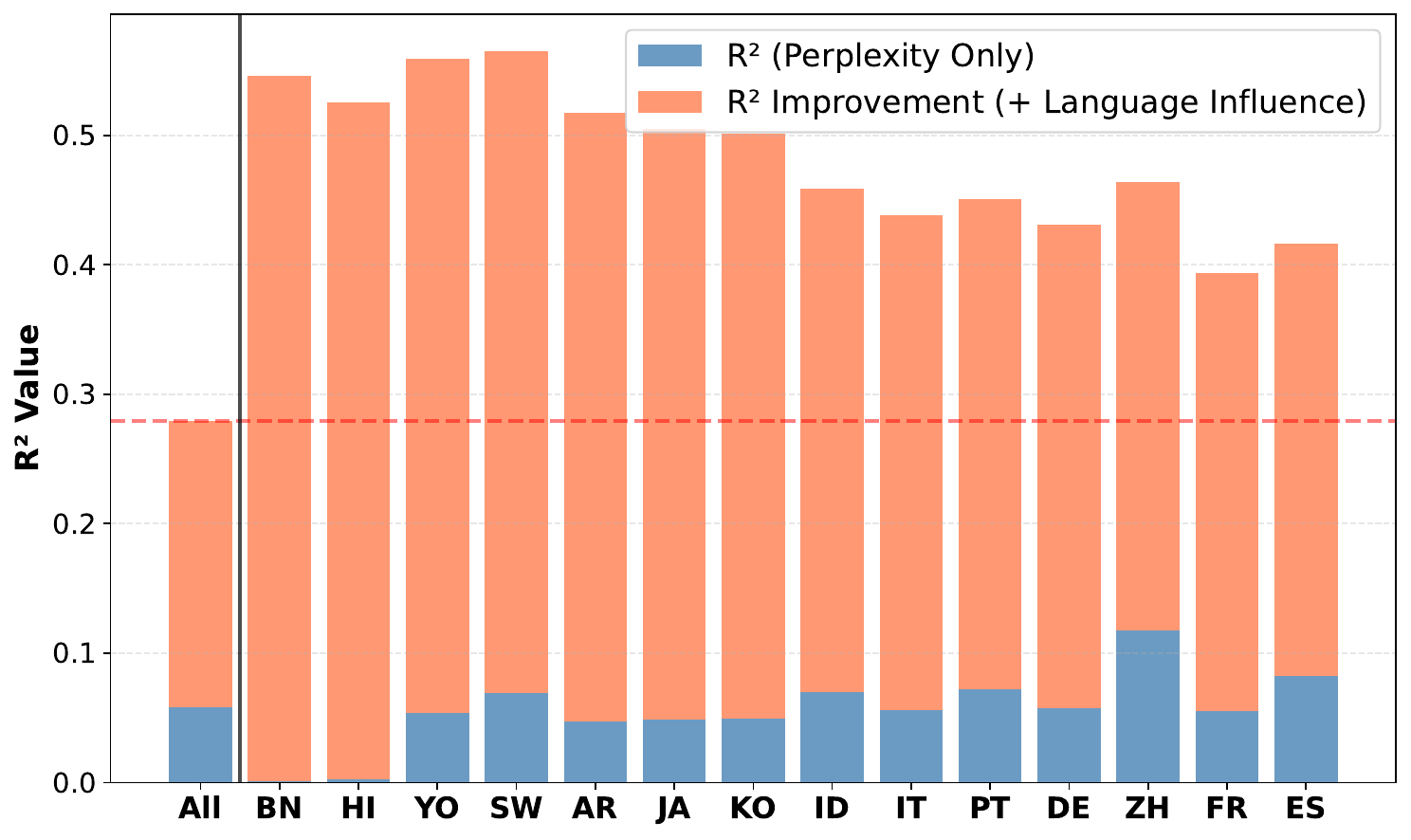}
  \caption{M-Prometheus-14B's $R^2$ decomposition: perplexity contribution (bottom) and additional language contribution (top) when first-language answers are incorrect and second-language answers are correct.}
  \label{fig:prometheus-14b-ft}
\end{figure}

\begin{figure}[t]
  \centering
  \includegraphics[width=\columnwidth]{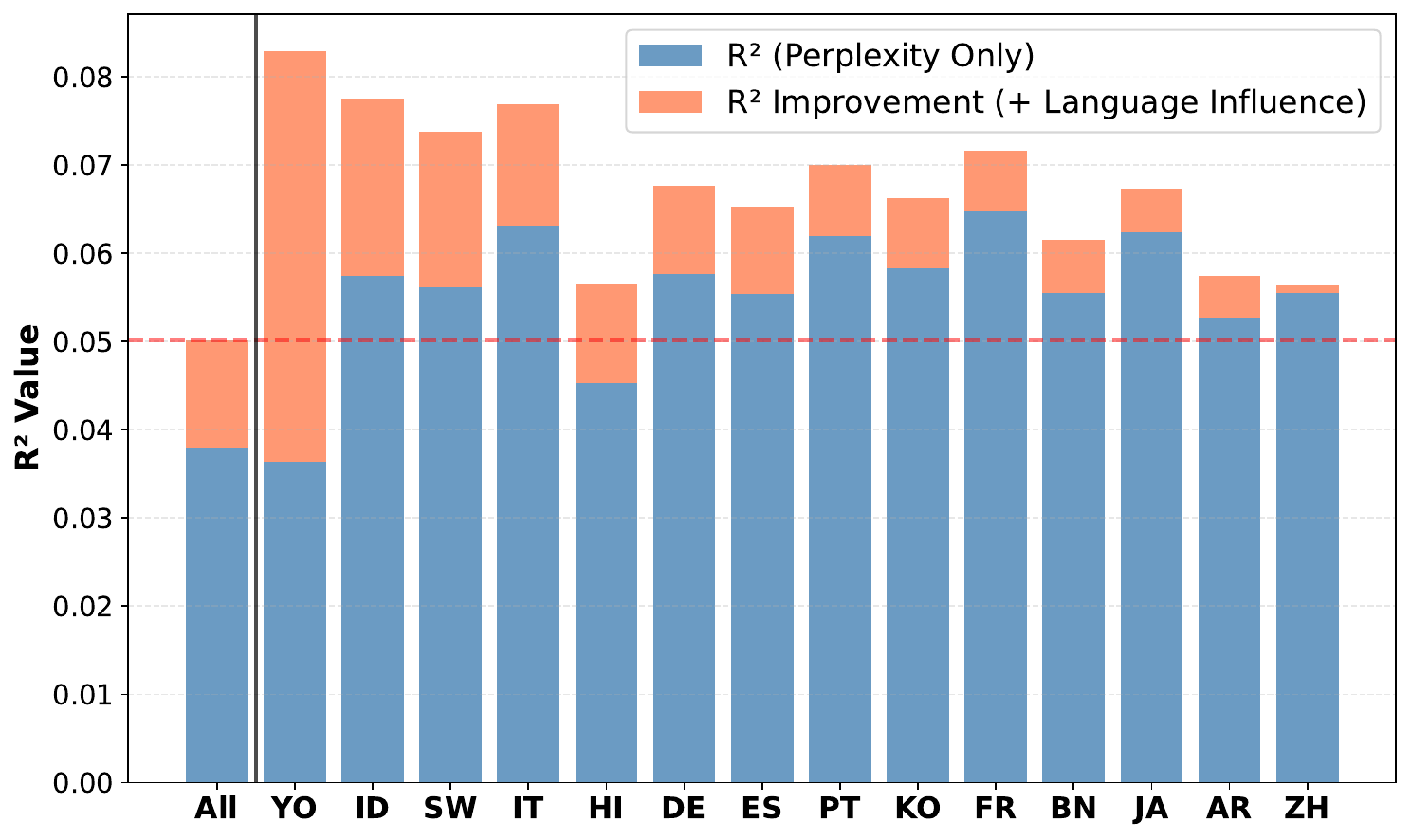}
  \caption{Llama 3.3 70B's $R^2$ decomposition: perplexity contribution (bottom) and additional language contribution (top) when first-language answers are correct and second-language answers are incorrect.}
  \label{fig:llama-tf}
\end{figure}

\begin{figure}[t]
  \centering
  \includegraphics[width=\columnwidth]{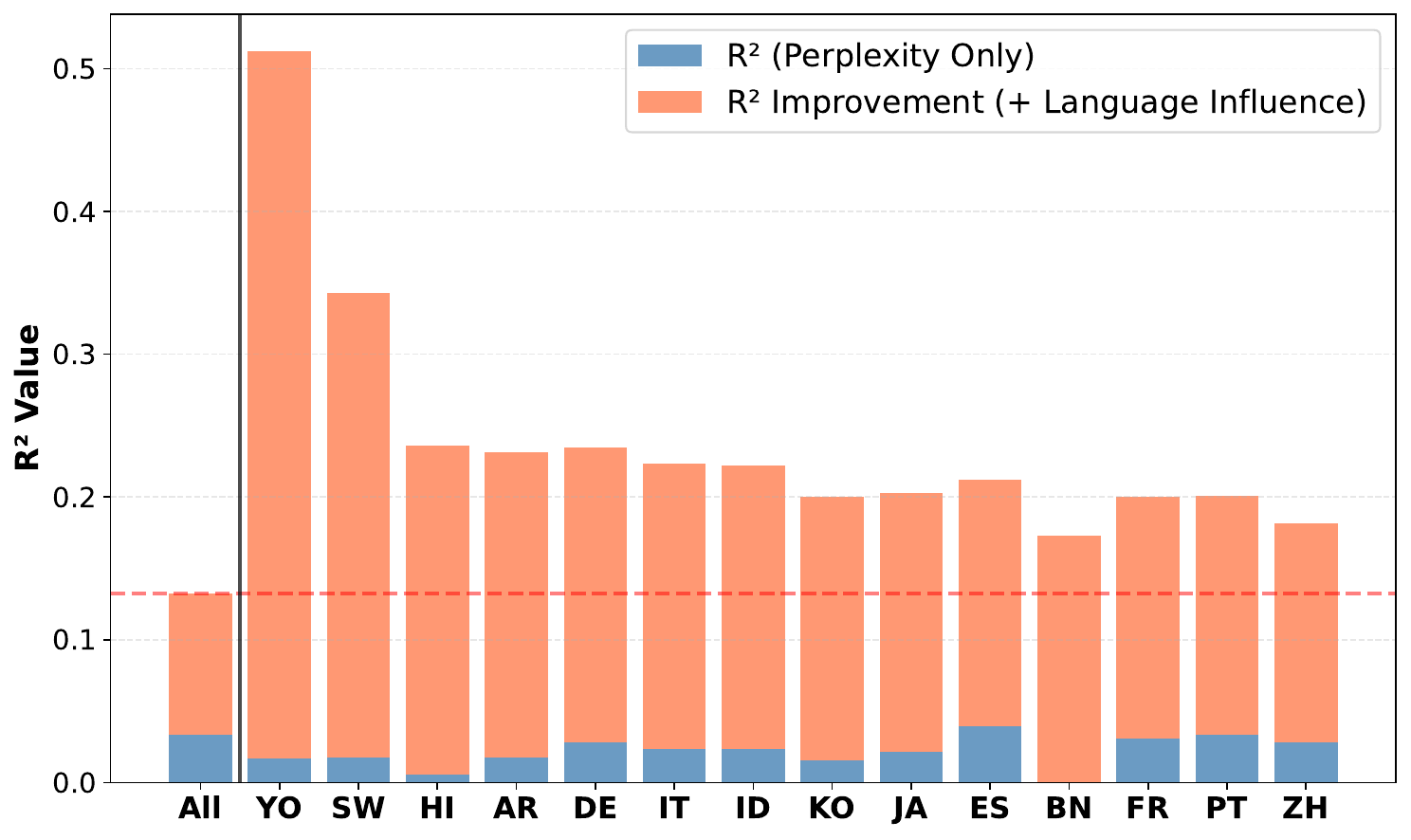}
  \caption{Llama 3.3 70B's $R^2$ decomposition: perplexity contribution (bottom) and additional language contribution (top) when first-language answers are incorrect and second-language answers are correct.}
  \label{fig:llama-ft}
\end{figure}

\begin{figure}[t]
  \centering
  \includegraphics[width=\columnwidth]{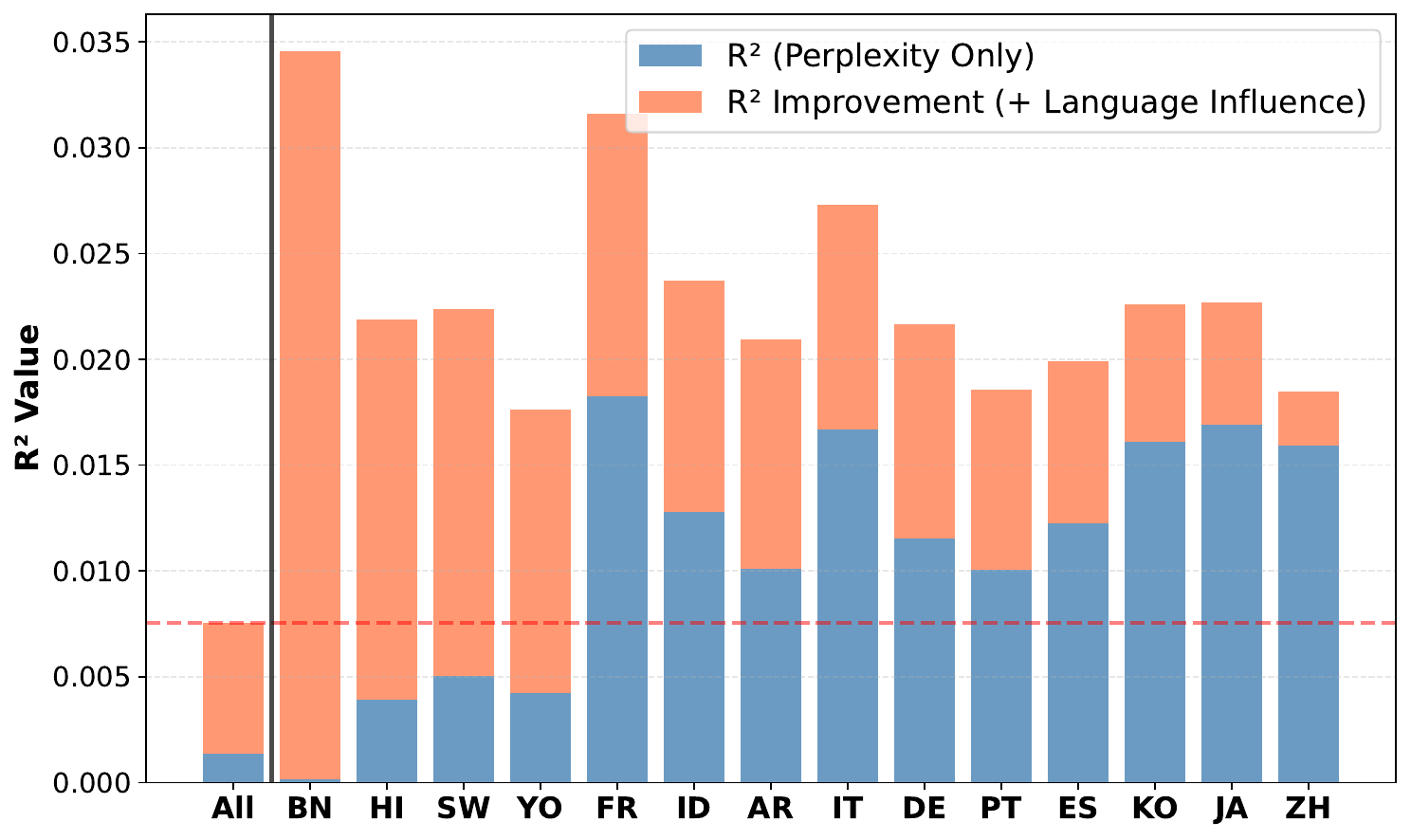}
  \caption{Aya-Expanse-32B's $R^2$ decomposition: perplexity contribution (bottom) and additional language contribution (top) when first-language answers are correct and second-language answers are incorrect.}
  \label{fig:aya-tf}
\end{figure}

\begin{figure}[t]
  \centering
  \includegraphics[width=\columnwidth]{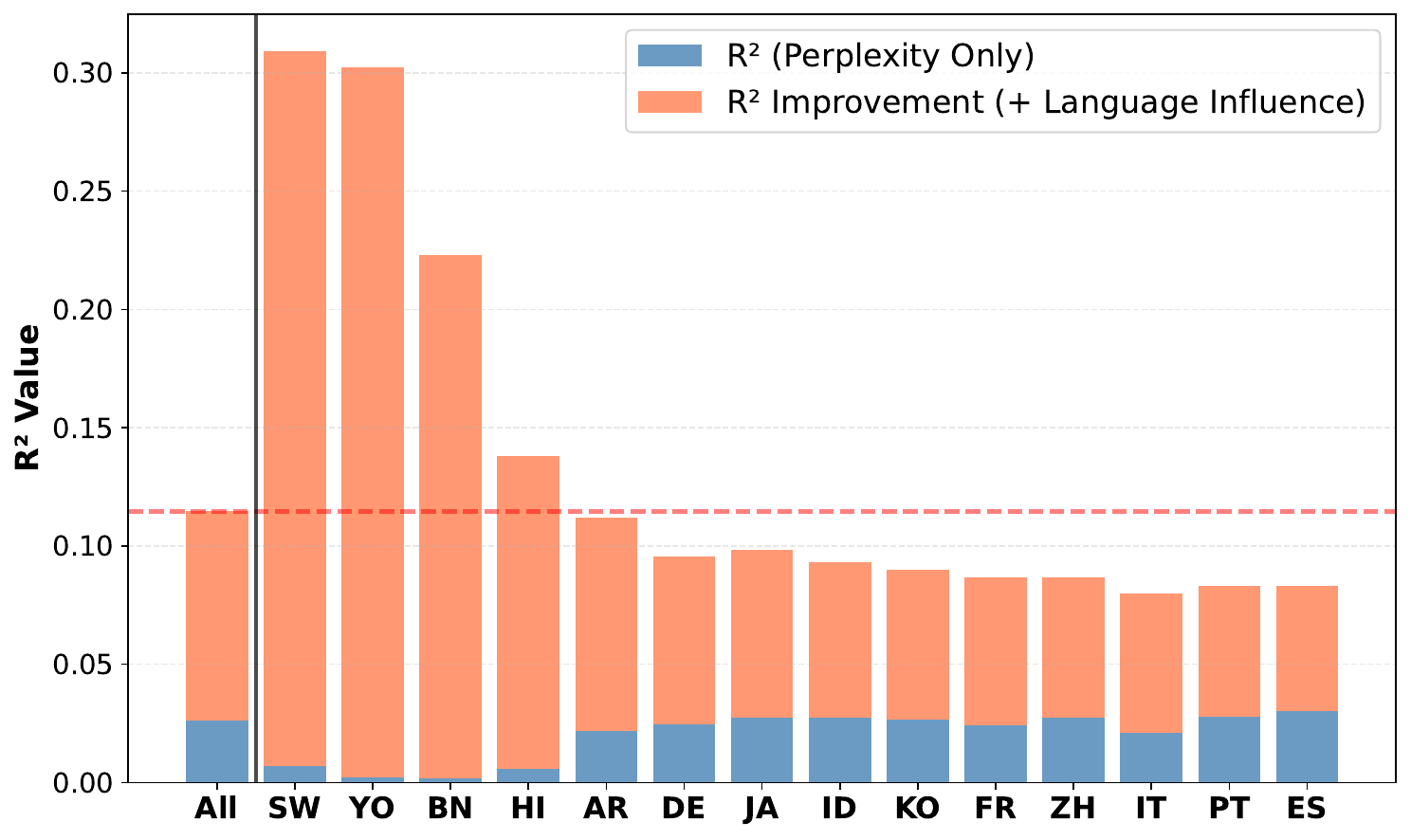}
  \caption{Aya-Expanse-32B's $R^2$ decomposition: perplexity contribution (bottom) and additional language contribution (top) when first-language answers are incorrect and second-language answers are correct.}
  \label{fig:aya-ft}
\end{figure}

\begin{figure}[t]
  \centering
  \includegraphics[width=\columnwidth]{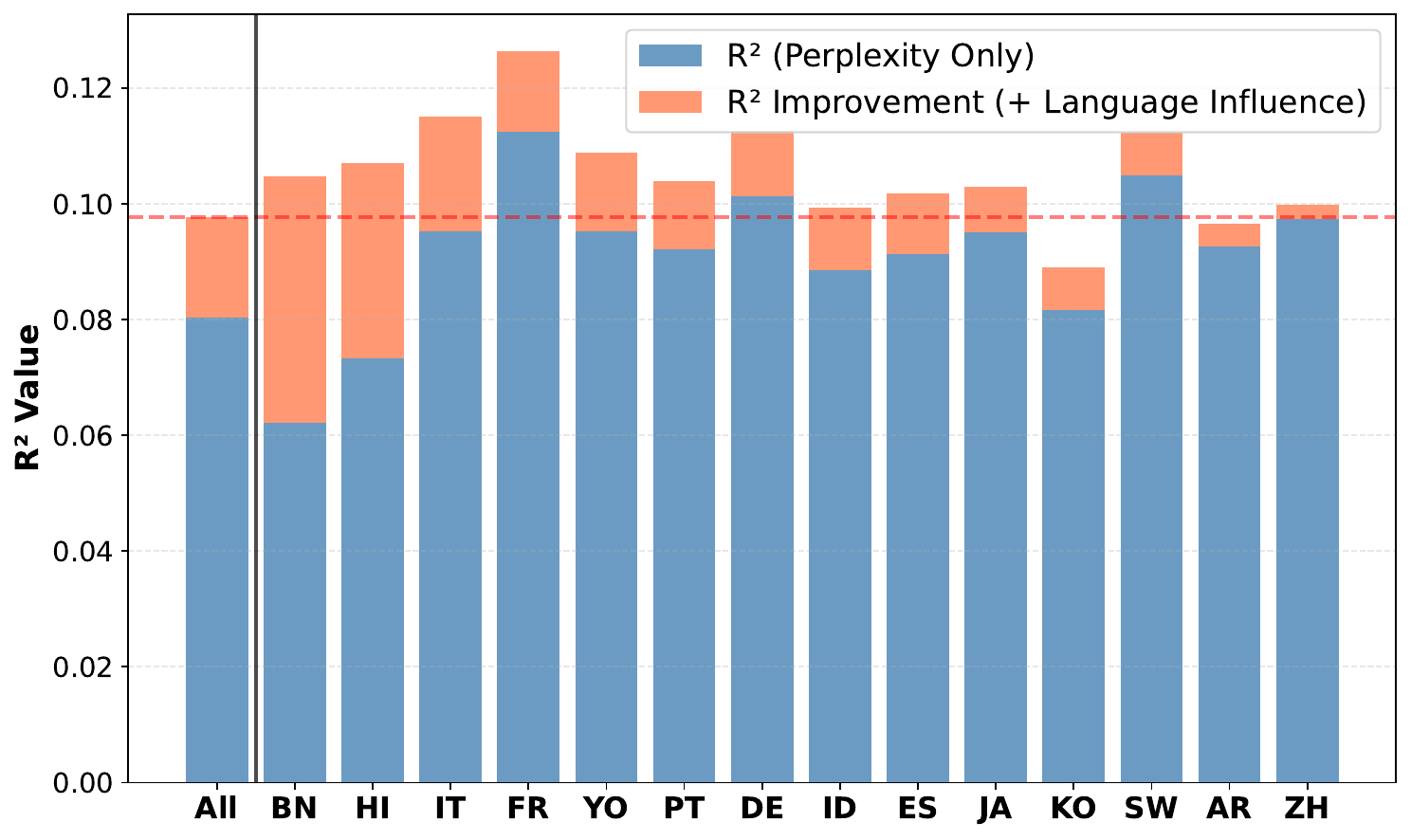}
  \caption{Qwen3-30B-A3B's $R^2$ decomposition: perplexity contribution (bottom) and additional language contribution (top) when first-language answers are correct and second-language answers are incorrect.}
  \label{fig:qwen-tf}
\end{figure}

\begin{figure}[t]
  \centering
  \includegraphics[width=\columnwidth]{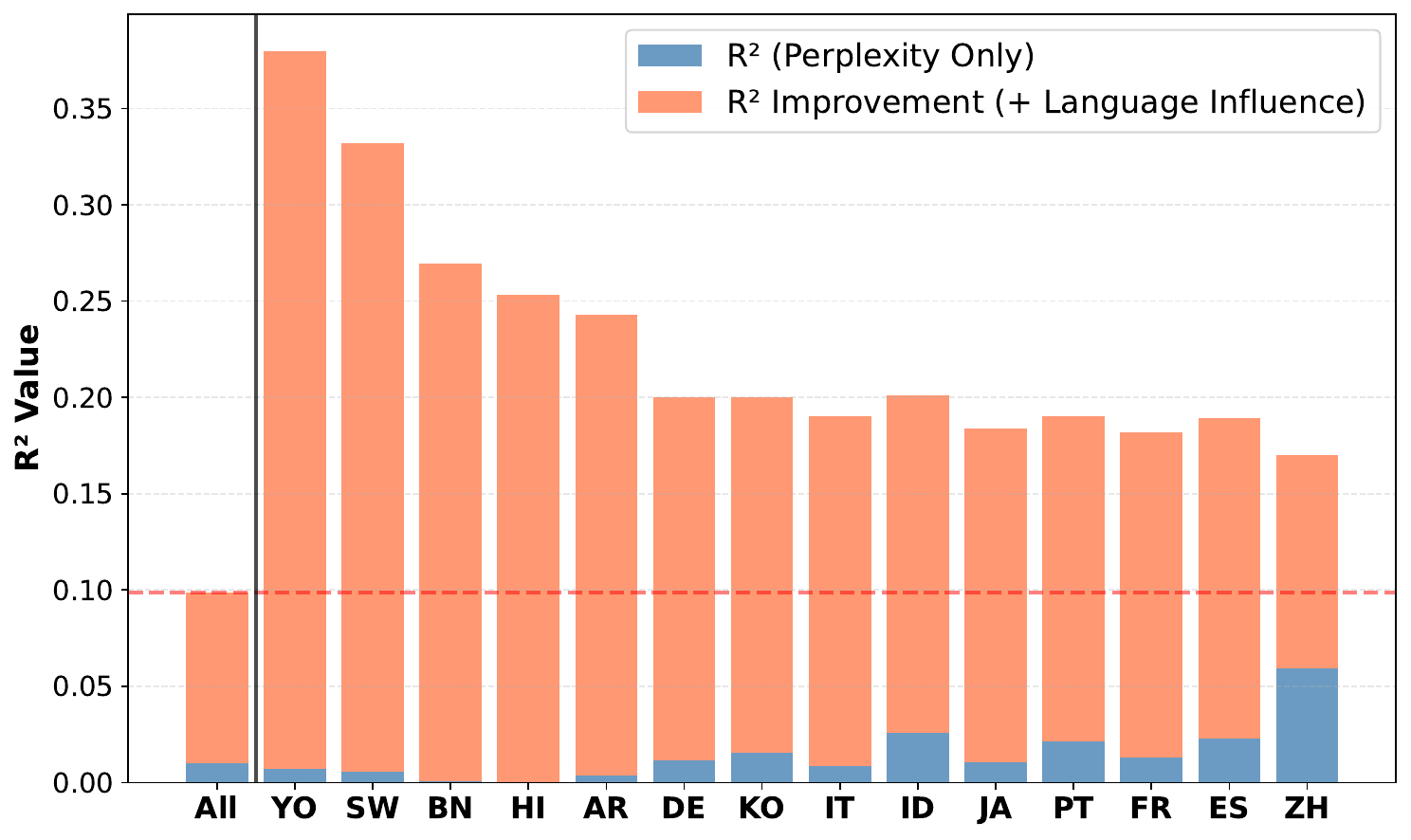}
  \caption{Qwen3-30B-A3B's $R^2$ decomposition: perplexity contribution (bottom) and additional language contribution (top) when first-language answers are incorrect and second-language answers are correct.}
  \label{fig:qwen-ft}
\end{figure}

The $R^2$ difference results for tested models can be seen in figure \ref{fig:prometheus-stacked_bar_tf}, \ref{fig:prometheus-14b-ft}, \ref{fig:llama-tf}, \ref{fig:llama-ft}, \ref{fig:aya-tf}, \ref{fig:aya-ft}, \ref{fig:qwen-tf} and \ref{fig:qwen-ft}.

\begin{figure*}[t]
  \centering
  \includegraphics[width=\textwidth]{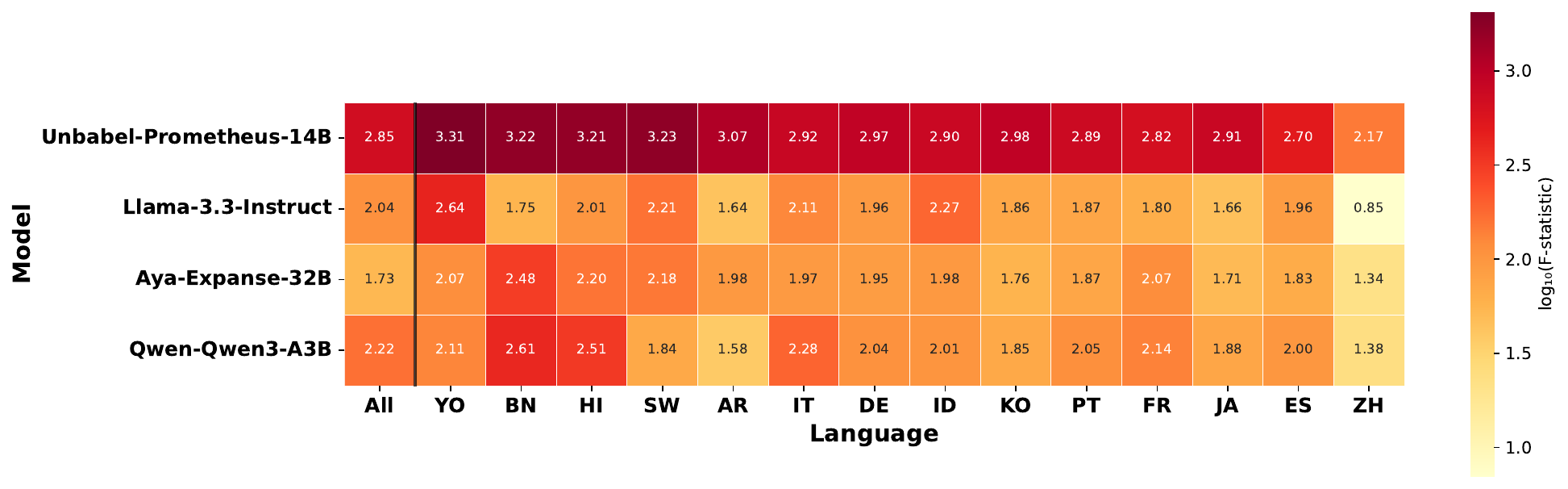}
  \caption{All models' F-statistics corresponding to each language when first-language answers are correct and second-language answers are incorrect.}
  \label{fig:heatmap-tf}
\end{figure*}

\begin{figure*}[t]
  \centering
  \includegraphics[width=\textwidth]{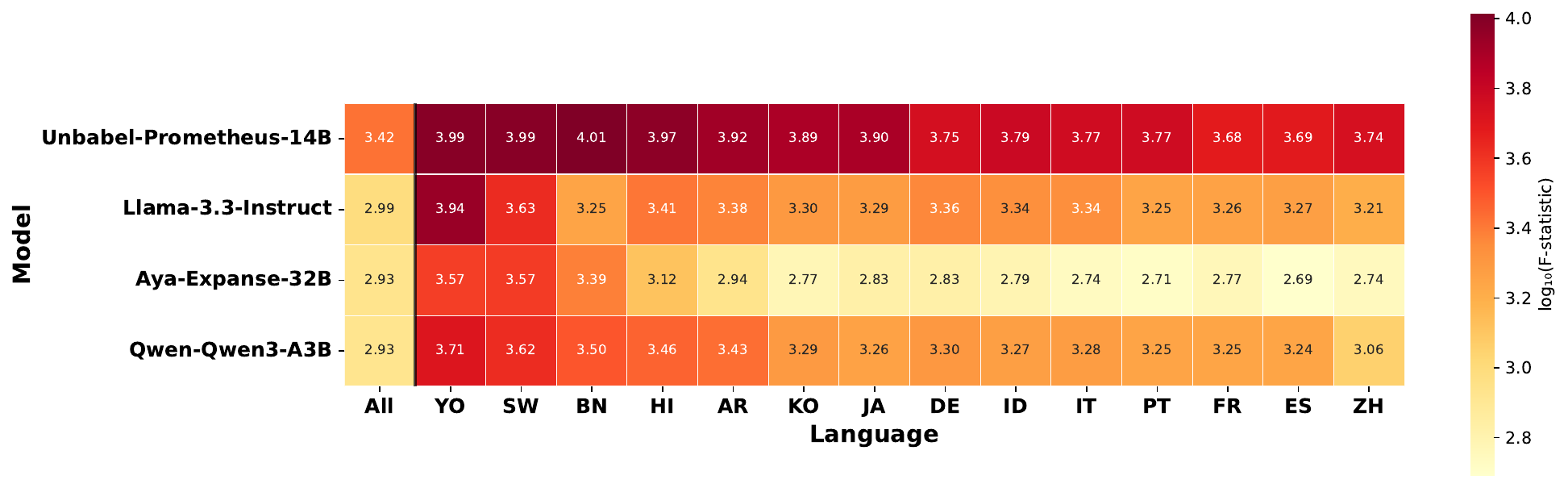}
  \caption{All models' F-statistics corresponding to each language when first-language answers are incorrect and second-language answers are correct.}
  \label{fig:heatmap-ft}
\end{figure*}

The F-statistic results for tested models can be seen in figure \ref{fig:heatmap-tf} and \ref{fig:heatmap-ft}.

\section{Prompt for Filtering Dataset for Perplexity}
Many of the questions in the original dataset are not suitable for collecting perplexity because they are open-ended to an extent that it is impossible for the LLM to come up with the exact same answer as the ground truth. For example, the question ``Which of the following is true?" would greatly confuse an LLM, and instead of giving a random true statement, it is more likely to ask for choices or clarification, so the perplexity of answering one of the choices will be very large and not meaningful. To identify and filter out such question and answer pairs, we use GPT-5 and give it the following prompt:

\begin{promptbox}[Prompt for Filtering Dataset for Perplexity]
\textbf{[System]}\\
You are a helpful assistant. 
The user will provide a multiple choice question, 4 answer choices and the correct answer. 
Your task is to judge whether there is a reasonably high chance for a sufficiently intelligent being 
to output the exact correct answer among the four choices, even if the choices are not given.
Questions that fail to meet this requirement may include:

- Questions that contain 'Which of the following...' or similar phrasing. This makes it impossible to hit the correct answer without seeing the choices. For example, 'Which of the following has a red color?' A: apple. B:...

- Questions that are open-ended to an extent that the correct answer cannot be hit with a reasonable chance without seeing the choices. For example, Shakespeare has the motto \_\_\_. A: To thine own self be true. B:...

Please note that if a question has a fixed answer but with multiple possible phrasings, it is still valid. For example, What was the most important finding by the House of Lords in the Pinochet case? A: The Pinochet case confirmed that all public acts enjoy immunity. B:...

Your output should only contain either 'VALID' or 'INVALID', without any additional explanation.

\textbf{[User]}

Question: [question]

Answer1: [answer1]

Answer2: [answer2]

Answer3: [answer3]

Answer4: [answer4]

Correct answer: [correct answer]
\end{promptbox}

\section{AI Usage Statement}

During the completion of this project, we used LLMs to 1) summarize and facilitate understanding of related works, 2) accelerate coding and debugging, and 3) polish our writing and check for grammar errors. The core ideas of this paper are our own, and we have double-checked all LLM-generated outputs to make sure they are consistent with our intentions.

\end{document}